\documentclass[conference]{IEEEtran}
\usepackage{times}

\pdfoutput=1

\usepackage{multicol}
\usepackage{color}
\usepackage{colortbl}  
\usepackage{array} 
\usepackage{float}
\usepackage{chemformula}
\usepackage{url}
\usepackage{algorithm}  
\usepackage{algorithmicx} 
\usepackage{algpseudocode}
\usepackage{adjustbox}

\usepackage{ragged2e}
\usepackage{wrapfig}
\usepackage[accsupp]{axessibility}
\usepackage{amsfonts}

\usepackage{marvosym}

\usepackage[numbers]{natbib}                            
\usepackage{booktabs}                                   
\usepackage{multirow}                                   
\usepackage{makecell}                                   
\usepackage{tablefootnote}                              
\usepackage[symbol]{footmisc}                           
\usepackage{amsmath,amssymb}                            
\usepackage{xcolor}                                     
\let\labelindent\relax                                  
\usepackage{enumitem}                                   
\usepackage{subcaption}                                 
\usepackage{stfloats}                                   

\usepackage[bookmarks=true]{hyperref}                   
\usepackage{graphicx} 

\usepackage{epsfig}
\usepackage{epstopdf} 

\usepackage{cleveref}
\usepackage{csquotes}
\usepackage{xspace}

\usepackage{cite}
\usepackage{nicematrix}
\usepackage{bm}
\usepackage{pifont}%
\newcommand{\cmark}{\ding{51}}%
\newcommand{\xmark}{\ding{55}}%

\newcommand{\dgray}[1]{\textcolor{darkgray}{#1}}
\newcommand{\gray}[1]{\textcolor{gray}{#1}}
\newcommand{\orange}[1]{\textcolor{orange}{#1}}

\newcommand{\bbluecell}[0]{\cellcolor[HTML]{E0F4FF}}

\newcommand{\eat}[1]{}                                  

\newcommand{\ours}[0]{{RoboMIND}\xspace}
\newcommand{\ntrajs}{107k}
\newcommand{\ntasks}{479}
\newcommand{\nobjs}{96}
\newcommand{\nskills}{38}
\newcommand{\nhours}{305.5}

\newcommand{\humantk}[2]{#2}                             

\newcommand{\nrobot}{Tien Kung\xspace}

\begin{document}

\title{\ours: Benchmark on Multi-embodiment Intelligence Normative Data for Robot Manipulation}

\author{%
 Kun Wu$^{1,*}$, Chengkai Hou$^{2,3,*}$, Jiaming Liu$^{2,3,*}$, Zhengping Che$^{1,*,\dagger}$, Xiaozhu Ju$^{1,*,\dagger}$,\\
 Zhuqin Yang$^{1}$, Meng Li$^{1}$, Yinuo Zhao$^{1}$, Zhiyuan Xu$^{1}$, Guang Yang$^{1}$, Shichao Fan$^{1}$, Xinhua Wang$^{1}$, Fei Liao$^{1}$,\\
 Zhen Zhao$^{1}$, Guangyu Li$^{1}$, Zhao Jin$^{1}$, Lecheng Wang$^{1}$, Jilei Mao$^{1}$, Ning Liu$^{1}$, Pei Ren$^{1}$, Qiang Zhang$^{1}$,\\
 Yaoxu Lyu$^{2}$, Mengzhen Liu$^{2,3}$, Jingyang He$^{2,3}$, Yulin Luo$^{2,3}$, Zeyu Gao$^{3}$, Chenxuan Li$^{2}$, Chenyang Gu$^{2,3}$,\\ 
 Yankai Fu$^{2}$, Di Wu$^{2}$, Xingyu Wang$^{2}$, Sixiang Chen$^{2,3}$, Zhenyu Wang$^{2,3}$, Pengju An$^{2,3}$, Siyuan Qian$^{2,3}$,\\
 Shanghang Zhang$^{2,3,\text{\Letter}}$, Jian Tang$^{1,\text{\Letter}}$\\
 $^1$Beijing Innovation Center of Humanoid Robotics\\
 $^2$State Key Laboratory of Multimedia Information Processing, School of Computer Science, Peking University\\
 $^3$Beijing Academy of Artificial Intelligence
}


\twocolumn[
{%
\renewcommand\twocolumn[1][]{#1}
\maketitle
\begin{center}
  \centering
  \begin{minipage}[t]{\linewidth}
    \includegraphics[width=0.99\textwidth]{\humantk{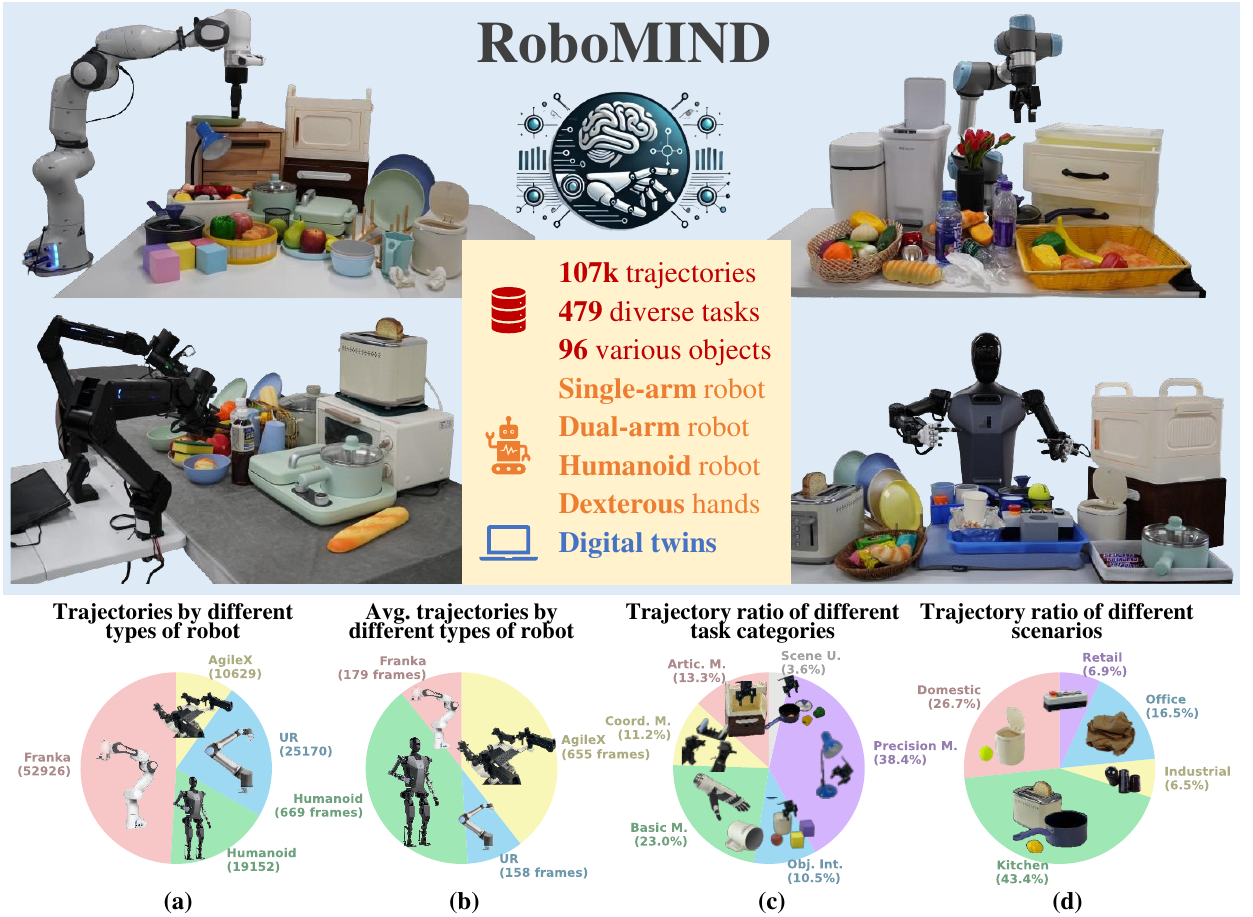}{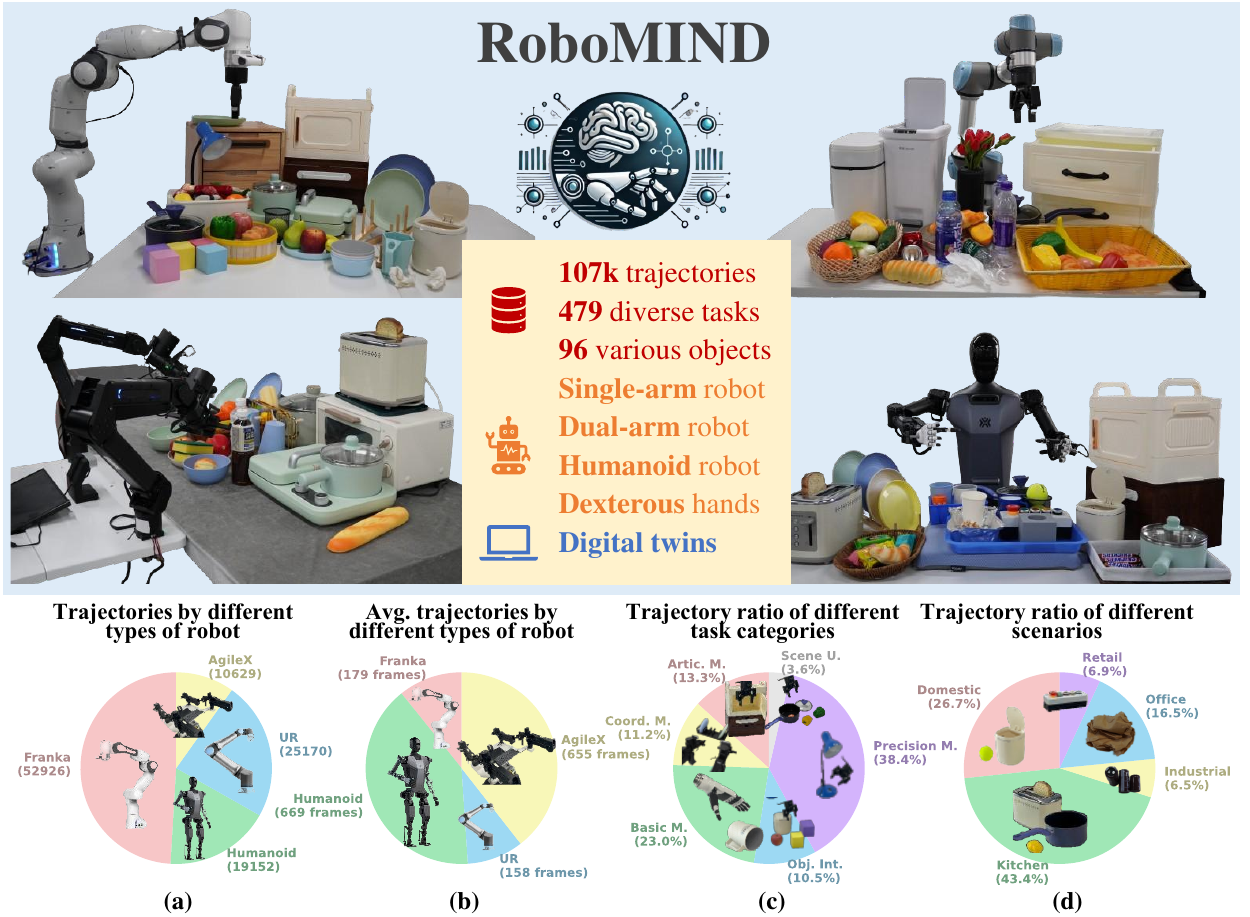}}
    {\captionsetup{hypcap=false}  
    \captionof{figure}{\footnotesize{\textbf{
    Overview of \ours. 
    We introduce \ours (Multi-embodiment Intelligence Normative Data for Robot Manipulation), 
    comprising {\ntrajs} demonstration trajectories across {\ntasks} diverse tasks involving {\nobjs} distinct object classes. 
    To ensure consistency and reliability during policy learning, \ours is gathered through human teleoperation and structured around a unified data collection standard.
    The four pie charts represent 
    (a) the total trajectory numbers categorized by different types of robots, 
    (b) average trajectory lengths (frames) categorized by different types of robots, 
    (c) trajectory ratio of different task categories 
    (Artic. M.: Articulated Manipulations; Coord. M.: Coordination Manipulations; Basic M.: Basic Manipulations; Obj. Int.: Multiple Object Interactions; Precision M.: Precision Manipulations; Scene U.: Scene Understanding), 
    and (d) trajectory ratio of different scenarios.
    }}}
    \label{fig:teaser}}
  \end{minipage}
\end{center}
}]

\begin{abstract}
Developing robust and general-purpose manipulation policies is a key goal in robotics.
To achieve effective generalization, it is essential to construct comprehensive datasets that encompass a large number of demonstration trajectories and diverse tasks.
Unlike vision or language data, which can be sourced from the internet, robotic datasets require detailed observations and manipulation actions, necessitating significant investments in both hardware-software infrastructure and human labor.
While existing works have focused on assembling various individual robot datasets, there is still a lack of a unified data collection standard and insufficient high-quality data across diverse tasks, scenarios, and robot types.
In this paper, we introduce \ours (Multi-embodiment Intelligence Normative Data for Robot Manipulation), a dataset containing {\ntrajs} demonstration trajectories across {\ntasks} diverse tasks involving {\nobjs} object classes. 
\ours is collected through human teleoperation and encompasses comprehensive robotic-related information, including multi-view observations, proprioceptive robot state information, and linguistic task descriptions.
To ensure data consistency and reliability for imitation learning, \ours is built on a unified data collection platform and a standardized protocol, covering four distinct robotic embodiments: 
the Franka Emika Panda, \humantk{a}{the X-Humannoid \nrobot} humanoid robot with dual dexterous hands, the AgileX dual-arm robot, and the UR5e.
Our dataset also includes 5k real-world failure demonstrations, each accompanied by detailed causes, enabling failure reflection and correction during policy learning.
Additionally, we created a digital twin environment in the Isaac Sim simulator, replicating the real-world tasks and assets, which facilitates the low-cost collection of additional training data and enables efficient evaluation. 
To demonstrate the quality and diversity of our dataset, we conducted extensive experiments using various imitation learning methods for single-task settings and state-of-the-art Vision-Language-Action (VLA) models for multi-task scenarios. 
By leveraging \ours, the VLA models achieved high manipulation success rates and demonstrated strong generalization capabilities.
To the best of our knowledge, \ours is the largest multi-embodiment teleoperation dataset collected on a unified platform, providing large-scale and high-quality robotic training data.
Our project is at \href{https://x-humanoid-robomind.github.io/}{https://x-humanoid-robomind.github.io/}.

\end{abstract}


\footnotetext[0]{%
 \\
 $^{1}$Beijing Innovation Center of Humanoid Robotics, Beijing, China
 \textit{\{Gongda.Wu, z.che, jason.ju, jian.tang\}@x-humanoid.com}%
 \\ 
 $^{2}$State Key Laboratory of Multimedia Information Processing, School of Computer Science, Peking University, Beijing, China
 \textit{shanghang@pku.edu.cn}%
 \\
 $^{*}$Co-first authors: Kun Wu, Chengkai Hou, Jiaming Liu, Zhengping Che, and Xiaozhu Ju
 \\
 $^{\dagger}$Project leaders: Zhengping Che and Xiaozhu Ju
 \\
 $^{\text{\Letter}}$Corresponding authors: Shanghang Zhang and Jian Tang
}

\section{Introduction}

One of the aspirations of any professional in the field of robotics is to develop a versatile, general-purpose robotic model capable of performing a broad spectrum of real-world tasks.
Specifically, such models should be generalizable in order to execute the intended manipulation tasks under varying conditions, such as a new robot, unfamiliar environments, or different objects~\citep{o2023open,khazatsky2024droid,kim2024openvla,liu2024rdt,liu2024robomamba,cheang2024gr}. 
To achieve this level of generalization, researchers have drawn inspiration from the training of large models in computer vision and natural language processing, where rich and diverse datasets have proven essential~\citep{achiam2023gpt,lin2023videollava,team2023gemini,wang2023cogvlm,datacomp2024,li2024datacomplm}.
They concluded that for training generalizable robotic models, one of the most critical elements is the access to rich and diverse training data that encompass varied scenes, tasks, and robot types.
Such diversity ensures that models learn to perform reliably under different conditions and environments~\citep{mandlekar2018roboturk,o2023open,shafiullah2023bringing,cheang2024gr,fang2024rh20t,team2024octo}.
Therefore, in this work, we aim to \textbf{\textit{construct comprehensive datasets that capture a broad spectrum of robotic interactions and experiences to facilitate training models capable of mastering various manipulation policies.}}

However, the curation of large-scale datasets for training general-purpose robotic models poses significant challenges.
In contrast to the acquisition of vision or language data, which can often be sourced through web-based collection methods~\citep{datacomp2024,li2024datacomplm}, 
collecting robotic data is difficult because such data cannot be easily obtained in the same way, as it requires controlled environments where the joints and end-effector information of robotic systems are meticulously recorded.
Moreover, scaling up data collection efforts necessitates considerable investment in both hardware and software infrastructure and human labor for oversight, particularly when it comes to acquiring and curating high-quality demonstration data~\citep{o2023open,wang2024all,khazatsky2024droid}.
Consequently, even the most versatile robotic manipulation policies currently in use are predominantly trained on datasets gathered within constrained conditions that offer limited diversity in robot types~\citep{o2023open,khazatsky2024droid}. 

Our dataset, called \ours (\textbf{M}ulti-embodiment \textbf{I}ntelligence \textbf{N}ormative \textbf{D}ata for \textbf{Robo}t manipulation), is an extensive dataset that encompasses a broad range of robotic interactions and experiences.
{\ours} features \textbf{\ntrajs} demonstration trajectories amounting to {\nhours} hours of interaction data of 4 kinds of robotic embodiments including Franka Emika Panda~\citep{franka_site}, a humanoid robot\humantk{}{ (i.e., X-Humanoid \nrobot~\citep{tien_kung_site})},
AgileX Cobot Magic V2.0~\citep{agilex_site}, and UR5e~\citep{ur5e_site}, as shown in Figure~\ref{fig:teaser}. 
Unlike the Open X-Embodiment dataset~\citep{o2023open}, which was compiled from various laboratories with differing data collection standards and diverse combinations of robotic platforms, 
\ours is gathered within the same standardized setting, adhering to a standardized data collection protocol to ensure consistency and reliability. 
By maintaining uniform data collection standards, all data points are captured under similar conditions, reducing variability and noise, which is crucial for training models that can generalize well across different tasks and environments.
The standardized procedures also enhance the reliability of the dataset, making it easier to validate and reproduce experimental results, thereby building trust in the trained models and ensuring their consistent performance in real-world applications.

Moreover, {\ours} covers a wide range of robot environments and spans {\ntasks} diverse tasks involving {\nobjs} various object classes. 
Additionally, we provide a dataset from our real-world tasks simulated in the Nvidia Isaac Sim~\citep{isaacsim}.
{\ours} incorporates data from various robot types, including 26,856 motion trajectories from Franka Emika Panda single-arm robots, 15,187 from \humantk{the}{\nrobot} humanoid robots, 10,269 from AgileX Cobot Magic V2.0 dual-arm robots, 25,170 from UR5e single-arm robots, and 30,035 from simulation. 
All these trajectories are collected through a teleoperation system that captures natural human motion patterns and maps them onto robots to drive the same motion trajectories. 
These trajectories encompass RGB-D data from distinct viewpoints, detailed proprioceptive state information of the robot body, specific information regarding the robot's end effector, and a linguistic description of the task at hand.  
Containing such comprehensive and detailed information, these data are valuable for training robotic models to perform complex manipulation tasks.

At the same time, we not only publish the {\ntrajs} successful trajectories but also document the 5k trajectories of real-world failure cases.
The robot model can explore the causes of failures by learning from these failure case trajectories, thereby improving its performance through such learning experiences.
This technique is representative of Reinforcement Learning from Human Feedback (RLHF)~\citep{christiano2017deep,ouyang2022training}, where human oversight and feedback direct the learning process of models, leading the models to produce more desirable and accurate outcomes.
In addition, we annotate a total of 10k robot trajectories in \ours with frame-level fine-grained language descriptions. 
These annotated trajectories encompass a wide range of robot tasks.
To ensure accuracy and reliability, each annotation undergoes verification and correction by multiple reviewers.
We believe that these additional failure cases and fine-grained linguistic annotations will further advance research in robot learning, particularly in areas such as failure recovery~\citep{liu2023reflect}, task planning~\citep{liang2023code}, visual question answering~\citep{driess2023palm}, among others.

\begin{table*}[t]
  \centering
  \setcounter{footnote}{1}
  \caption{
    Comparison to existing real-world datasets for robot manipulation. 
    All data is drawn from the original paper or from the DROID paper~\citep{khazatsky2024droid}.
    We divide robot types into three categories: single-arm, dual-arm, and humanoid. 
    We report the number of unique multi-view trajectories and highlight the advantages of \ours in \orange{orange}. 
    \small{
      \dgray{$^\ddag$non-robot, tool-based data collections.}
      \gray{$^\S$not a dataset in itself, but an \emph{aggregation} of existing datasets.}
    }
  }
  \resizebox{1.0\textwidth}{!}{
  \begin{tabular}{l c c c c c c c c c c l}
    \toprule
    \textbf{Dataset} & \textbf{Trajectory} & \textbf{Task}  & \textbf{Skill} & \textbf{Arm} & \textbf{Dexterous Hand}& \textbf{Detailed Annotation}  &  \textbf{Robot Type}  & \textbf{Public Robot} & \textbf{Failure Data} & \textbf{Digital Twin} & \textbf{Collection} \\
    \midrule
    \citet{pinto2016supersizing}                & 50k   & n/a & 1   & Dual   & \xmark & \xmark & 1 & \cmark & \cmark & \xmark & Scripted \\
    Home-LCA~\citep{gupta2018robot}             & 28k   & n/a & 1   & Single & \xmark & \xmark & 1 & \xmark & \xmark & \xmark & Scripted \\
    BrainRobotData~\citep{levine2018learning}   & 800k  & n/a & 1   & Single & \xmark & \xmark & 1 & \xmark & \cmark & \xmark & Scripted \\
    Roboturk~\citep{mandlekar2018roboturk}      & 2.1k  & 3   & 2   & Single & \xmark & \xmark & 1 & \xmark & \cmark & \xmark & Human Teleoperation \\
    MIME~\citep{sharma2018multiple}             & 8.2k  & 20  & 20  & Single+Dual & \xmark & \xmark & 1 & \xmark & \xmark & \xmark & Human Teleoperation \\
    Sketchy~\citep{cabi2019scaling}             & 74.4k & 5   & n/a & Single & \xmark & \cmark & 1 & \cmark & \cmark & \xmark & 12\% Human / 78\% Scripted \\
    RoboNet~\citep{dasari2020robonet}           & 162k  & n/a & n/a & Single & \xmark & \xmark & 1 & \cmark & \xmark & \xmark & Scripted \\
    BridgeData~\citep{ebert2021bridge}          & 7.2k  & 71  & 4   & Single & \xmark & \xmark & 1 & \cmark & \xmark & \xmark & Human Teleoperation \\
    MT-Opt~\citep{kalashnikov2021mt}            & 800k  & 12  & 1   & Single & \xmark & \xmark & 1 & \cmark & \xmark & \xmark & Scripted \\
    RT-1~\citep{brohan2022rt}                   & 130k  & 700 & 8   & Single & \xmark & \xmark & 1 & \xmark & \xmark & \xmark & Human Teleoperation \\
    BC-Z~\citep{jang2022bc}                     & 26k   & 100 & 3   & Single & \xmark & \xmark & 1 & \xmark & \xmark & \xmark & Human Teleoperation \\
    BridgeData~V2~\citep{walke2023bridgedata}   & 60.1k & n/a & 13  & Single & \xmark & \xmark & 1 & \cmark & \xmark & \xmark & 85\% Human / 15\% Scripted \\
    RoboSet~\citep{bharadhwaj2024roboagent}     & 98.5k & 38  & 6   & Single & \xmark & \xmark & 1 & \cmark & \xmark & \xmark & 30\% Human / 70\% Scripted \\
    RH20T~\citep{fang2024rh20t}                 & 13k   & 140 & 33  & Single & \xmark & \xmark & 1 & \cmark & \xmark & \xmark & Human Teleoperation \\
    DROID~\citep{khazatsky2024droid}            & 76k   & n/a & 86  & Single & \xmark & \xmark & 1 & \cmark & \xmark & \xmark & Human Teleoperation \\
    BRMData~\citep{zhang2024empowering}         & 0.5k  & 10  & 7   & Dual   & \xmark & \xmark & 1 & \cmark & \xmark & \xmark & Human Teleoperation \\
    \dgray{Dobb-E~\citep{shafiullah2023bringing}$^\ddag$} & \dgray{5.6k} & \dgray{109} & \dgray{6}  & \dgray{Single}       & \dgray{\xmark}  & \dgray{\xmark}  & \dgray{1}  & \dgray{\cmark} & \dgray{\xmark}  & \dgray{\xmark}  & \dgray{Human Tool-based}   \\
    \gray{Open~X-Embodiment~\citep{o2023open}$^\S$}       & \gray{1.4M}  & \gray{160k} & \gray{217} & \gray{Single+Dual}   & \gray{\xmark}   & \gray{\xmark}   & \gray{2}   & \gray{\cmark}  & \gray{\xmark}   & \gray{\xmark}   & \gray{Dataset Aggregation} \\
    \cmidrule(lr){1-12}
    \ours                                                 & \ntrajs      & \ntasks     & \nskills   & \orange{Single+Dual} & \orange{\cmark} & \orange{\cmark} & \orange{3} & \cmark         & \orange{\cmark} & \orange{\cmark} & Human Teleoperation        \\
    \bottomrule
  \end{tabular}}
  \setcounter{footnote}{0}
  \captionsetup{width=\textwidth}
  \label{tab:dataset_cmp}
\end{table*}

Beyond establishing such a large-scale and diverse dataset, we conduct extensive experiments to not only validate the dataset's effectiveness but also evaluate various algorithms' performance, providing a comprehensive benchmark analysis.
Specifically, we evaluate the task success rates using single-task imitation learning methods, including ACT~\citep{zhao2023learning}, Diffusion Policy~\citep{chi2023diffusion_policy}, and BAKU~\citep{haldar2024baku}. Additionally, we assess the generalization capabilities and task success rates of Vision-Language-Action (VLA) large models such as OpenVLA~\citep{o2023open}, RDT-1B~\citep{liu2024rdt}, and CrossFormer~\citep{doshi2024scaling}. 
The experimental results demonstrate that \ours can be effectively utilized by various single-task imitation learning algorithms and successfully adapted to VLA large models. 
The high-quality information provided by our dataset enables successful task execution across different approaches in real-world scenarios.
Furthermore, pre-training the entire VLA models using the full \ours dataset results in significant improvements in task performance across multiple robot types. 
To simplify the use of \ours, we provide the code scripts that adapt \ours files with the open-source LeRobot framework~\citep{cadene2024lerobot} at \href{https://github.com/x-humanoid-robomind/x-humanoid-training-toolchain/}{https://github.com/x-humanoid-robomind/x-humanoid-training-toolchain/}. 

\section{Related Work}

\textbf{Robotic Manipulation.}
Traditional manipulation policies typically rely on state-based reinforcement learning~\citep{andrychowicz2020learning,joshi2020robotic,yarats2021mastering}. 
In contrast, recent works~\citep{mo2021where2act,eisner2022flowbot3d,fang2023anygrasp} incorporate visual observations as input to predict action poses.
Imitation learning policies, in particular, enable robots to acquire stable manipulation skills by imitating an expert through demonstration~\citep{deng2018learning,wu2024swbt,zare2024survey}.
Driven by advancements in diffusion-based generative models~\citep{ho2020ddpm,song2020ddim,rombach2022stable_diffusion}, diffusion policy~\citep{chi2023diffusion_policy} and subsequent works~\citep{pearce2023imitating_diffusion,reuss2023goal,wu2024discrete} focus on transforming random Gaussian noise into coherent action sequences, with methods such as DP3~\citep{ze20243d} and 3D Diffuser Actor~\citep{ke20243d} further enhancing this process in 3D space. 
On the other hand, some Multimodal Large Language Models (MLLMs)~\citep{ahn2022can,driess2023palm,huang2023voxposer} enable robots to comprehend natural language and visual scenes, automatically generating task plans.
Meanwhile, Vision-Language-Action (VLA) models~\citep{zitkovich2023rt,li2023vision,li2023manipllm,liu2024robomamba,kim2024openvla} empower MLLMs to predict low-level $SE(3)$ poses, demonstrating interpretability and generalization in diverse scenarios.
Given the critical role of 3D spatial information in complex manipulation tasks, several works~\citep{zhao2023learning,goyal2023rvt,shridhar2023perceiver,gervet2023act3d} explore the encoding of point cloud data or multi-view images for 3D imitation learning. 
However, most existing methods are trained on simulation datasets or self-collected real-world datasets, and the robotics community still lacks a unified large-scale dataset.

\textbf{Robotic Learning Datasets.}
Interacting with spatial configurations in real-world environments is vital for robots. 
However, collecting data with a real robotic arm often incurs substantial costs~\citep{o2023open,khazatsky2024droid}.
General-purpose simulators~\citep{coumans2016pybullet,gazebo,isaacgym,isaacsim} replicate the physical world and provide virtual environments for training policy models, significantly reducing the costs and time associated with data collection.
To meet the training demands of complex and long-horizon tasks, simulators based on real-world environments are developed~\citep{AI2-THOR,chang2017matterport3d,SAPIEN,habitat}, featuring photorealistic 3D assets and scenes built with game engines. 
However, the sim-to-real gap significantly impacts the manipulation accuracy of imitation learning policies.
As a result, some research shifts towards directly collecting real-world data, including datasets gathered through automated scripts or expert agents~\citep{pinto2016supersizing,gupta2018robot,levine2018learning,cabi2019scaling,dasari2020robonet,kalashnikov2021mt}, as well as those obtained via human teleoperation~\citep{mandlekar2018roboturk,sharma2018multiple,ebert2021bridge,brohan2022rt,jang2022bc,walke2023bridgedata,bharadhwaj2024roboagent,fang2024rh20t}.
As shown in Table~\ref{tab:dataset_cmp}, we compare \ours with representative publicly available real-world datasets for robot manipulation. 
RoboSet~\citep{bharadhwaj2024roboagent} and BridgeData~V2~\citep{walke2023bridgedata} include over 50k trajectories, but are limited to 6 and 13 skill types, respectively.
In contrast, RH20T~\citep{fang2024rh20t} covers 33 tasks, 
while its data scale is relatively small compared to the others.
Recently, Open~X-Embodiment~\citep{o2023open} has made a large effort to unify existing robot datasets into a standardized format, incorporating data from diverse robots collected through collaboration among 21 institutions. 
Following this, ARIO~\citep{wang2024all} further integrates real-world and simulated data into a standard format, aiming to bridge the gaps in existing data resources. 
DROID~\citep{khazatsky2024droid} collects 76k demonstration trajectories via human teleoperation.
Although previous large-scale datasets offer diverse scenarios, most focus on a single embodiment type—the two-finger gripper—and lack dexterous hands, limiting task variety.
In contrast, our proposed \ours features four distinct embodiments, including both grippers and dexterous hands, and expands the number of task types to {\ntasks} with long-horizon dual-arm tasks for complex skill training. 
Most importantly, {\ours} is collected in a standardized setting, ensuring consistency and minimizing variability.

\textbf{Large-scale Policy Learning.}
Learning robotic policies from large and diverse datasets has become a major research focus in the field of robotics.
One series of works leverages egocentric human videos~\citep{goyal2017something,damen2018scaling,damen2022rescaling,grauman2022ego4d} to assist in robot action learning.
Leveraging large-scale human videos, previous works investigate learning robotic representations~\citep{nair2022r3m,bahl2023affordances}, manipulation priors~\citep{mandikal2022dexvip,kannan2023deft}, and dexterous hand control~\citep{mandikal2021learning,wu2023learning}.
Another prominent approach, VLA models, leverages multimodal instruction datasets~\citep{mao2016generation,liu2023visual,2023_url_ShareGPT} and robot data~\citep{brohan2022rt,mees2022calvin,sermanet2023robovqa,wen2024tinyvla} for co-training or pretraining, enhancing the model's reasoning and generalization abilities.
Specifically, RT-2~\citep{zitkovich2023rt} innovatively incorporates large-scale internet data and low-level action data for co-finetuning;
RoboFlamingo~\citep{li2023vision} directly loads the pretrained parameters from OpenFlamingo~\citep{awadalla2023openflamingo} for visual instruction tuning;
RoboMamba~\citep{liu2024robomamba} utilizes high-level common sense and robotics-related reasoning data for co-training.
Finally, a series of works~\citep{liu2024rdt,kim2024openvla,li2024cogact} leverage large assembler datasets, such as Open~X-Embodiment and ARIO, for pre-training.
The large-scale pre-training significantly enhances the fine-tuning efficiency and generalization capability of policy models.
Our proposed real-world dataset and digital twin simulator provide a large-scale pretraining dataset and a high-quality fine-tuning dataset for policy learning in real-world applications, whose efficacy is demonstrated via abundant experiments.

\section{Dataset Collection and Processing}

\begin{figure*}[t]
    \centering
    \subfloat{\includegraphics[width=0.8\textwidth]{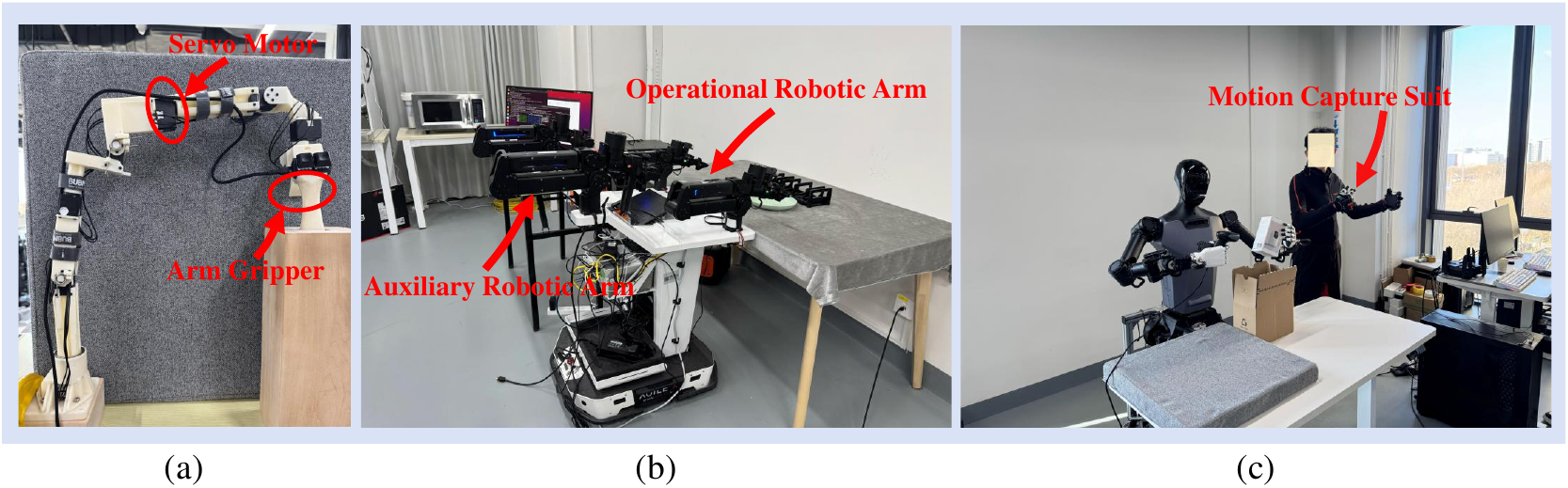}}
    \caption{
      Visualization of teleoperation methods for different robots. 
      (a) Using 3D-printed components to control the single-arm robots. 
      (b) Regulating the main robotic arm from the auxiliary arm for dual-arm operation. 
      (c) Adopting a motion capture suit to map onto the humanoid robot for operation.
    }
    \label{fig:operation}
    \vspace{-1em}
\end{figure*}

In this work, we primarily introduce how the \ours dataset is collected on the robots and detail the process of cleaning the \ours dataset. 
Our dataset is collected from four different robotic embodiments (Franka Emika Panda~\citep{franka_site}, \humantk{a humanoid robot}{\nrobot~\citep{tien_kung_site}}, AgileX Cobot Magic V2.0~\citep{agilex_site}, and UR5e~\citep{ur5e_site}), totaling \textbf{\ntrajs} trajectories on \textbf{\ntasks} tasks, \textbf{\nobjs} different object classes, and \textbf{\nskills} operational skills. 
To support the development of such a large-scale dataset, we develop an intelligent data platform designed to collect, filter, and process the dataset efficiently.
This platform uses a cloud-native architecture and distributed computing to handle large-scale data processing, offering five main functionalities and their corresponding modules:

\begin{enumerate}
  \item \textbf{Data Collection:} 
    Collect data from four types of robots using teleoperation equipments and then automatically transmit the collected data to the data platform;
  
  \item \textbf{Data Storage:} 
    Package and store the collected dataset in a standardized H5 format, including both visual data of the robot's executed actions and robotic proprioceptive data of its movements;
  
  \item \textbf{Data Preprocessing:} 
    Filter the dataset based on predefined standards, evaluating task execution accuracy, motion trajectory smoothness, and the presence of occlusion or motion blur in the visual data;
  
  \item \textbf{Data Classification:} 
    Categorize the collected dataset by robot type and specific tasks performed;
  
  \item \textbf{Data Annotation:}
    Perform detailed linguistic annotations on the collected dataset. 
\end{enumerate}

\subsection{Data Collection and Storage}

Teleoperation is widely applied in the data collection processes for various types of robots~\citep{rakita2017motion,zhang2018deep,wu2019teleoperation,handa2020dexpilot,liu2022robot,wu2023gello,qin2023anyteleop,cheng2024open,wang2024dexcap}.
Different types of robots also have specific teleoperation devices for data collection.
For example, researchers typically use VR headsets and motion capture suits to collect humanoid robot motion data. 
They capture the state of human movements and map this motion onto the humanoid robot platform, enabling the robot to replicate these movements while simultaneously collecting a comprehensive dataset~\citep{cheng2024open,wang2024dexcap}.
\ours contains teleoperation data in real-world and simulation environments from various types of robots, such as single-arm robots (Franka Emika Panda~\citep{franka_site}, UR5e~\citep{ur5e_site}), dual-arm robots (AgileX Cobot Magic V2.0~\citep{agilex_site}), and humanoid robots\humantk{}{ (X-Humanoid \nrobot~\citep{tien_kung_site})}.

\textbf{For the single-arm robots},
following the Gello~\citep{wu2023gello}, we construct the 3D-printed components and the servo motors that match the Degrees of Freedom (DoF) of the robotic arm (see Figure~\ref{fig:operation}(a)).
The motion of these 3D-printed components is mapped to the robotic arm's movements, thereby driving the arm. 
Additionally, we use depth cameras to record the RGB-D information of the robotic arm movement and simultaneously receive the robot state of the robotic arm.

\textbf{For the dual-arm robots},
we directly utilize a bilateral teleoperation device similar to the Mobile ALOHA system~\citep{fu2024mobile} on the robot to collect the dataset.
Figure~\ref{fig:operation}(b) shows that we employ a teleoperation structure using an auxiliary robotic arm to control the main robotic arm.

\begin{wrapfigure}{r}{2.5cm}
  \vspace{-0.5em}
  \includegraphics[width=2.5cm]{\humantk{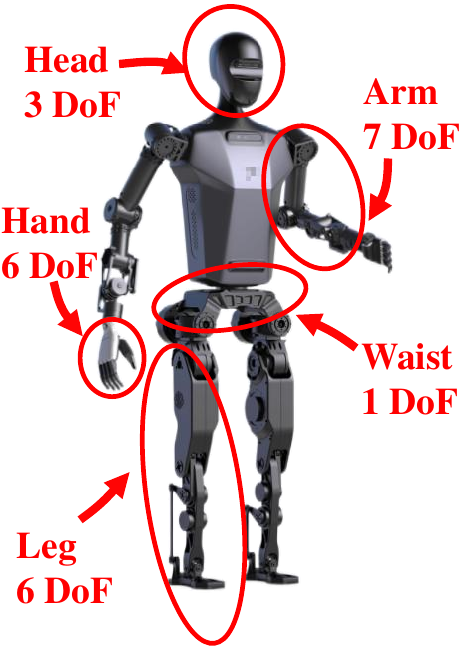}{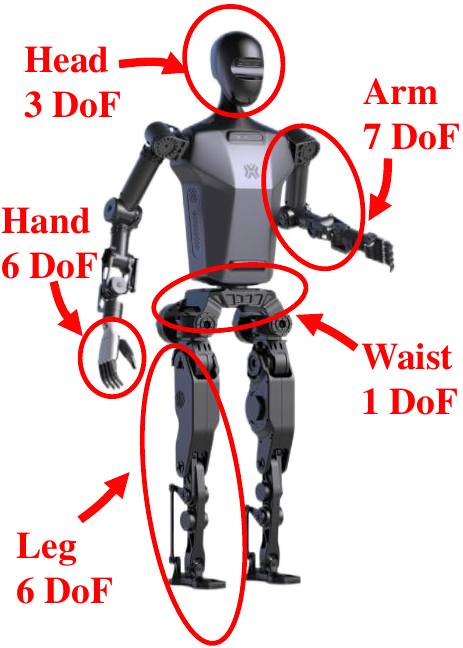}}
  \caption{\humantk{The}{The \nrobot} humanoid robot configuration.}
  \label{fig:ex}
  \vspace{-0.5em}
\end{wrapfigure}
\textbf{For the humanoid robots}, 
Figure~\ref{fig:ex} illustrates the structural design of \humantk{the}{the \nrobot} humanoid robot utilized in \ours.
In terms of configuration, it is highly modeled after humans.
The robotic arm is flexible and has a strong load-carrying capacity, making it suitable for performing operational tasks to collect datasets. 
The dexterous hand is integrated with multiple sensors for precise operation. 
With 42 degrees of freedom throughout the whole body, it can perform a wide variety of movements. 
In terms of visual perception, depth cameras are installed on its head, chest, waist, and back.
The head is equipped with the Orbbec Gemini 335~\citep{orbbec_gemini_335}, and the other parts are equipped with the Orbbec Gemini 335L~\citep{orbbec_gemini_335l}. 
These cameras use active and passive stereo vision technology to provide multiple data streams, accurately record visual perception information.
Besides Gello-style teleoperation devices, we use motion capture suits Xsens~\citep{xsens2025} to collect motion data from various joints of the human body and then map the human joint movements to the corresponding joint movements of a humanoid robot.
This allows the humanoid robot to perform the same actions as the human body, enabling remote operation for data collection.
Using motion capture suits provides a more accurate and direct method for capturing human movement, compared to relying on VR headsets~\citep{cheng2024open} and cameras~\citep{fu2024humanplus} for human pose recognition. 
Figure~\ref{fig:operation}(c) visualizes how we use a motion capture suit to collect data for humanoid robot operation.

To optimize storage efficiency and facilitate dataset organization, we consolidate each collected trajectory, encompassing multi-view RGB-D data, robot proprioceptive state information, specific end-effector state information, and teleoperation body state information, into a single H5 format file.

\begin{table}[t]
\caption{Examples of the task definitions for Franka, AgileX, and \humantk{the humanoid}{\nrobot} robots.}
\label{tab:taskname}
\centering
\resizebox{0.98\columnwidth}{!}{
  \begin{tabular}{cl}
    \toprule
    \textbf{Task Name} & \textbf{Task Description} \\ 
    \midrule
    \multirow{2}*{\texttt{FR-PlaceBreadPlate}} 
    & The Franka single-arm robot grasps a  \\ 
    & piece of bread and places it on a plate. \\
    \cmidrule{1-2}
    \texttt{AX-PackBowl} & The AgileX robot packs the bowls. \\
    \cmidrule{1-2}
    \multirow{2}{*}{\shortstack{\texttt{HR-OpenDrawer}\\\texttt{LowerCabinet}}}
    & The \humantk{humanoid}{\nrobot} robot opens the bottom \\ 
    & drawer of the cabinet. \\
    \bottomrule
  \end{tabular}
}
\vspace{-1em}
\end{table}

\begin{figure*}[t]
  \centering
  \subfloat{\includegraphics[width=0.95\textwidth]{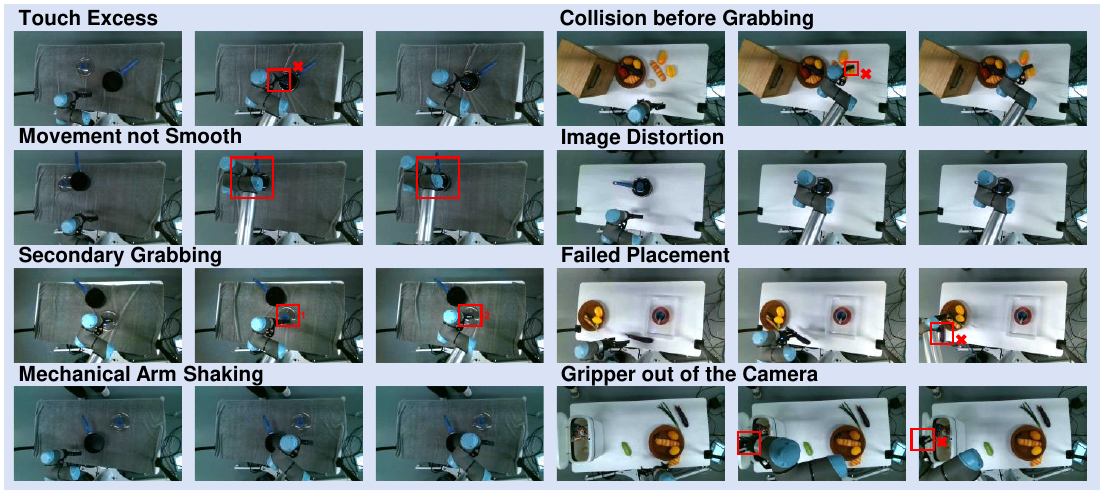}}
  \caption {
    We define 8 quality assurance criteria in the data collection process.
    \textbf{Touch Excess:} 
      Unnecessary contact with objects by the robotic arm; 
    \textbf{Movement not Smooth:} 
      Noticeable jerking or interruptions in robotic arm movements;
    \textbf{Secondary Grabbing:} 
      Repeated grasping attempts after failures in robotic arm operations;
    \textbf{Mechanical Arm Shaking:} 
      Abnormal vibrations in the robotic arm;
    \textbf{Collision before Grabbing:} 
      Collision of the gripper with surrounding objects before grasping;
    \textbf{Image Distortion:} 
      Data collection quality issues;
    \textbf{Failed Placement:} 
      Incorrect placement of objects;
    \textbf{Gripper out of the Camera:} 
      Frames in which the gripper exceeds video frame boundaries.
    During the data inspection process, all failures were annotated from videos.
    We show 8 trajectory examples that failed to pass the quality assurance due to different reasons.
    Each example includes three images that depict the dynamic process of the trajectory.
    We use \textcolor{red}{red boxes and markers} to highlight the reasons for failure.
  }
  \label{fig:quality_inspection}
  \vspace{-1em}
\end{figure*}

\subsection{Data Preprocessing and Classification}
All data is collected from operators controlling the teleoperation system in real-time, and errors can arise due to physical limitations such as fatigue, habits, distractions, or external disruptions.
To mitigate these issues, we employ a rotation rest system for operators and strive to provide a comfortable working environment to help them stay focused. 
Additionally, we perform comprehensive quality checks on collected data to ensure its reliability.
We define quality assurance criteria, such as avoiding unnecessary contacts and repeated grabbing (see Figure~\ref{fig:quality_inspection}). The quality assurance consists of three steps:
\begin{itemize}
  \item \textbf{Initial Inspection:}
    Quickly review videos to ensure there is no obvious technical issue, such as frame loss or freezing.
  \item \textbf{Detailed Inspection:}
    Review the video frame-by-frame or in slow motion to carefully check if the conditions described in Figure~\ref{fig:quality_inspection} are present.
  \item \textbf{Data Filtering and Issue Logging:}
    Document specific timestamps and descriptions for non-compliant data and categorize it for further processing or improvement.
\end{itemize}

For data classification, we adopt a task-centric data collection protocol, where each task serves as the fundamental unit of the dataset. 
We classify the collected datasets according to the task names, and each task name is comprehensively defined by four key components: 
(1) the specific robotic embodiment utilized; 
(2) the manipulation skill being executed;
(3) the objects involved in the task;
and (4) detailed scene descriptions, including object positions, spatial relationships, and environmental constraints or interfering elements. 
Table~\ref{tab:taskname} shows examples of the task definition.

This structured task-based framework ensures systematic data collection and enables fine-grained analysis of robotic manipulation capabilities across different scenarios and tasks.

\begin{figure*}[t]
  \centering
  \subfloat{\includegraphics[width=0.95\textwidth]{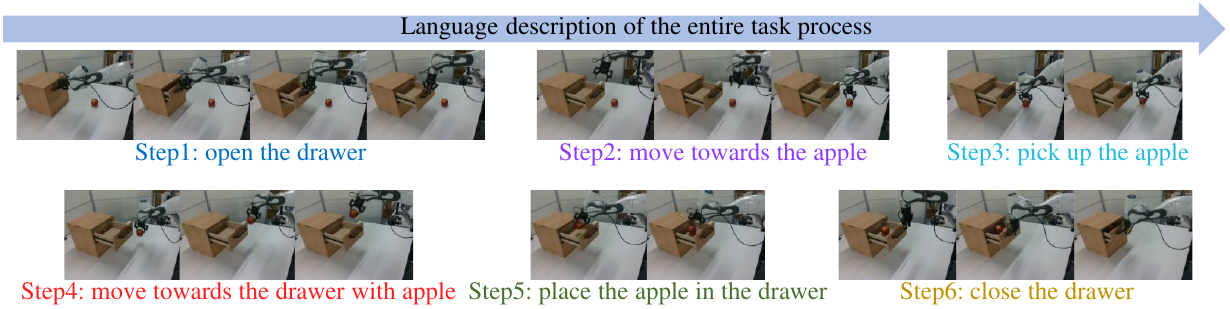}}
  \caption{
    Example of language description annotation.
    The video of the robotic arm placing the apple in the drawer is divided into six segments using Gemini.
    The language descriptions provided for each segment were initially generated by Gemini and subsequently refined through manual revision.
  }
  \label{fig:ln}
  \vspace{-1em}
\end{figure*}

\subsection{Data Annotation}
While the visual and robot proprioceptive information can be extracted directly from the collected videos and trajectories, we need to provide better semantic information from the data to aid model training.
For each collection task, its detailed and accurate linguistic descriptions are provided.
These linguistic annotations can be utilized for training currently popular VLA models.
In addition, \ours collection tasks encompass numerous long horizon tasks, where a uniform linguistic description may be insufficient to capture the full complexity and nuances of the entire task.
Thus, we offer detailed fine-grained linguistic annotations for each movement occurring within a trajectory, as illustrated in Figure~\ref{fig:ln}.
We annotate 10k successful robot motion trajectories, which are contained in long horizon manipulation tasks.
The annotation process involves two primary steps.
First, we use Gemini~\citep{team2023gemini} to segment each video based on the sequence of operations and generate detailed text descriptions for each segment. 
These descriptions accurately capture the operational steps and relevant context.
Second, we manually refine Gemini’s annotations regarding the following key aspects:
\begin{itemize}
    \item Identifying key manipulated objects;
    \item Detecting and describing all critical actions in the video;
    \item Ensuring accurate description of operational details;
    \item Applying reasonable granularity in temporal segmentation;
    \item Maintaining consistent temporal logic.
\end{itemize}

This thorough process enhances the precision and reliability of the language annotations for the collected trajectories. 
We show the annotation of a video of a Franka Emika Panda arm picking the apple and placing it in the drawer using the above standard procedure in Figure~\ref{fig:ln}.
The results show that our annotation scheme can accurately segment the key actions in the video and provide precise language descriptions of these key actions.
These detailed descriptions can be used for training models like RT-H~\citep{rth2024arxiv}.

\section{Dataset Analysis}

Based on a standardized procedure, we collected a large-scale, multi-embodiment dataset named \ours. 
This dataset consists of {\ntrajs} high-quality trajectories across 4 robotic embodiments, {\ntasks} tasks, {\nobjs} object classes, and {\nskills} skills. 
Robotic data diversity plays a crucial role in model generalization, encompassing various dimensions across hardware and environmental settings.
In this section, we perform a thorough quantitative analysis of key diversity dimensions, including robot variety, task length variation, task diversity, and object diversity. 
We analyze \ours across these dimensions, showing that it offers comprehensive training data to learn generalizable manipulation policies.
Furthermore, unlike previous works~\citep{o2023open,khazatsky2024droid}, \ours offers unique data types, such as language descriptions and failure case demonstrations, which enhance the policy model's ability to perform fine-grained task planning and reflect on failure actions.

\begin{figure}[tb]
  \centering
  \subfloat[
    Skill number distribution histogram for each embodiment. 
    We observe that over 70\% of the Franka tasks involve only a single skill, while over 75\% of the \humantk{humanoid}{\nrobot} and AgileX tasks involve two or more skills, indicating that these dual-arm tasks are mostly long-horizon tasks.
  ]
  {
    \includegraphics[width=0.9\columnwidth]{
      \humantk{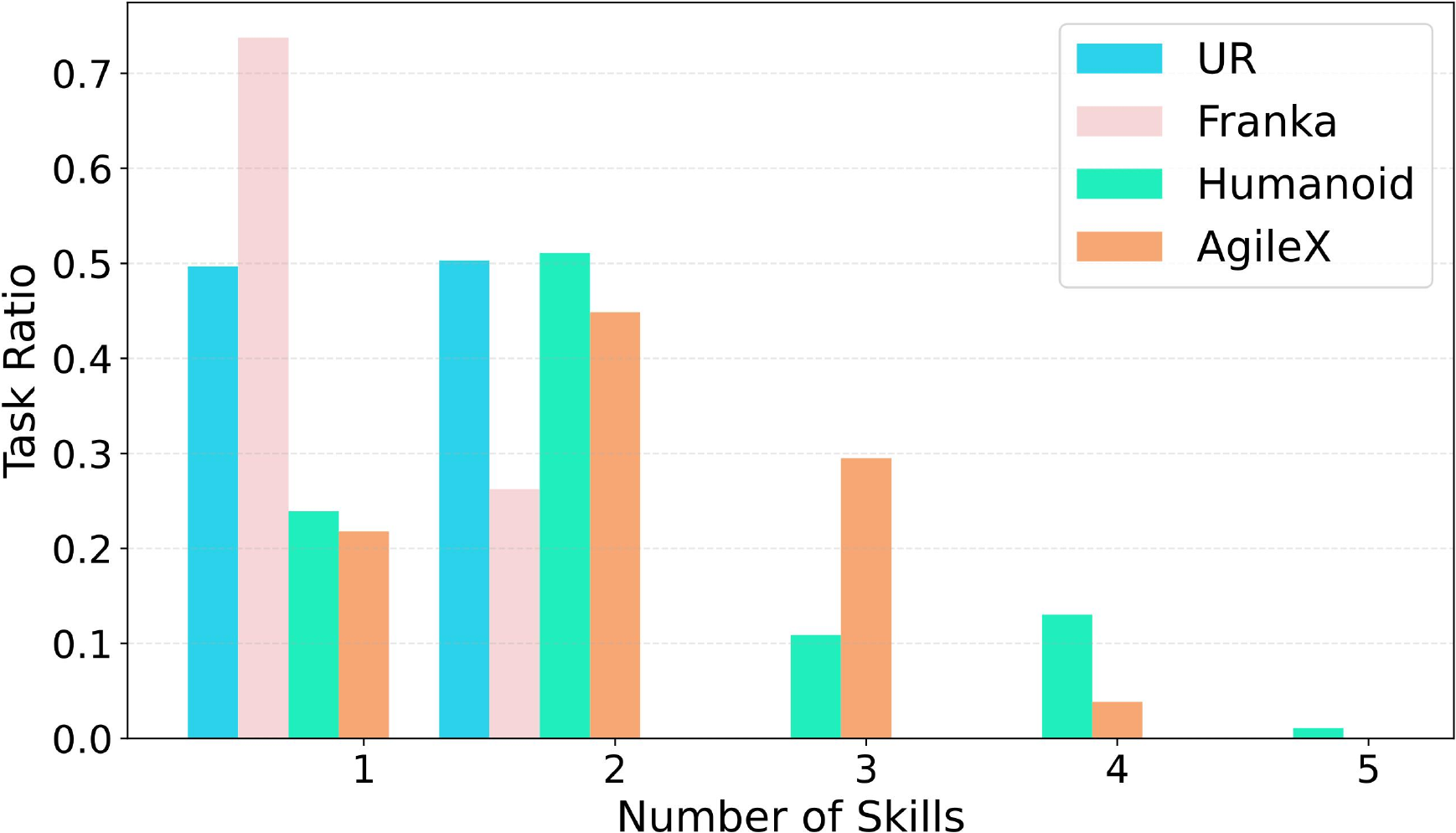}{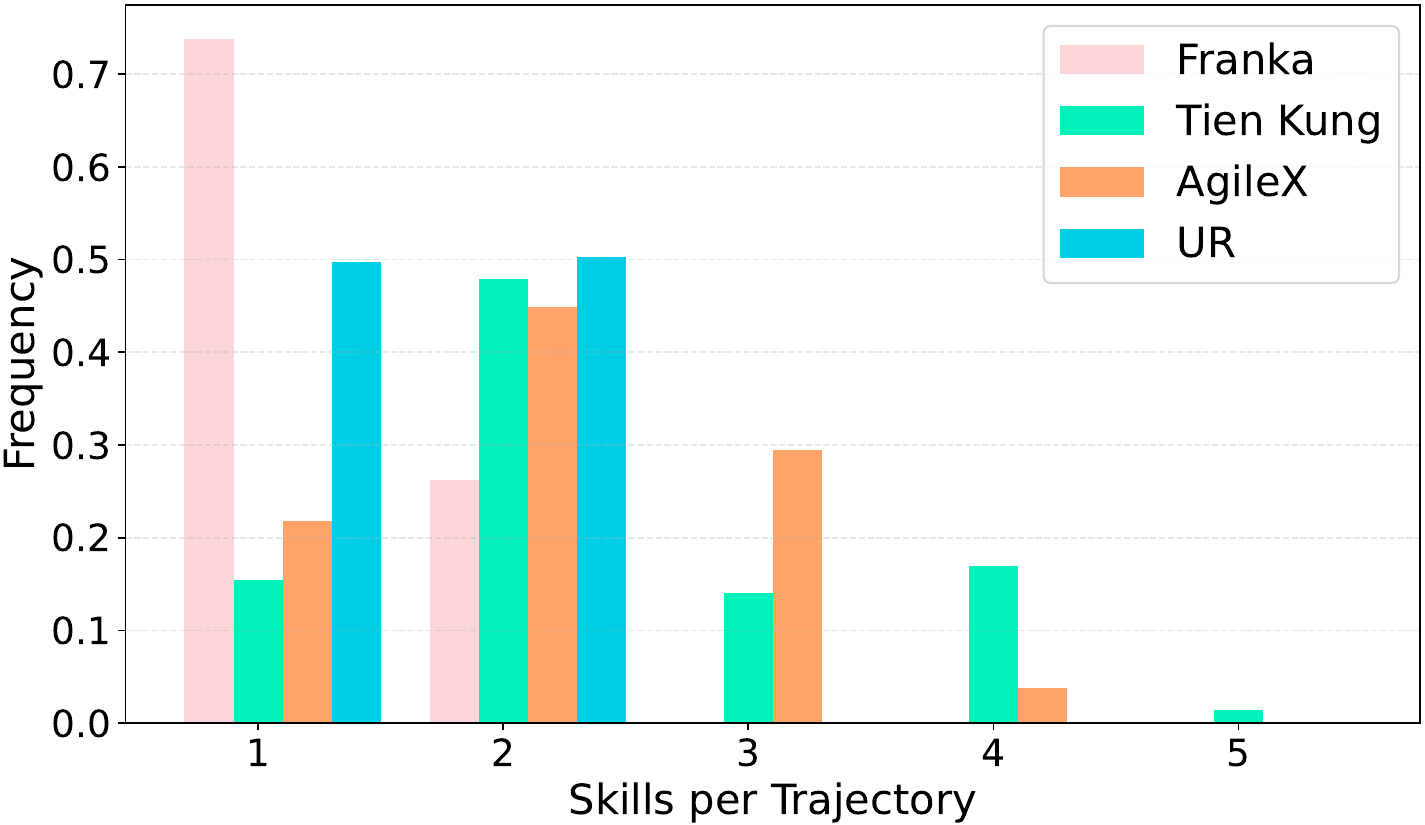}
    }
    \label{fig:skill_hist}
  }
  
  \centering
  \subfloat[
    The \texttt{AX-PutCarrot} task with the AgileX robot is visualized, involving a sequence of three different skills: pick, hand over, and place.
  ]
  {
    \includegraphics[width=0.9\columnwidth]{
      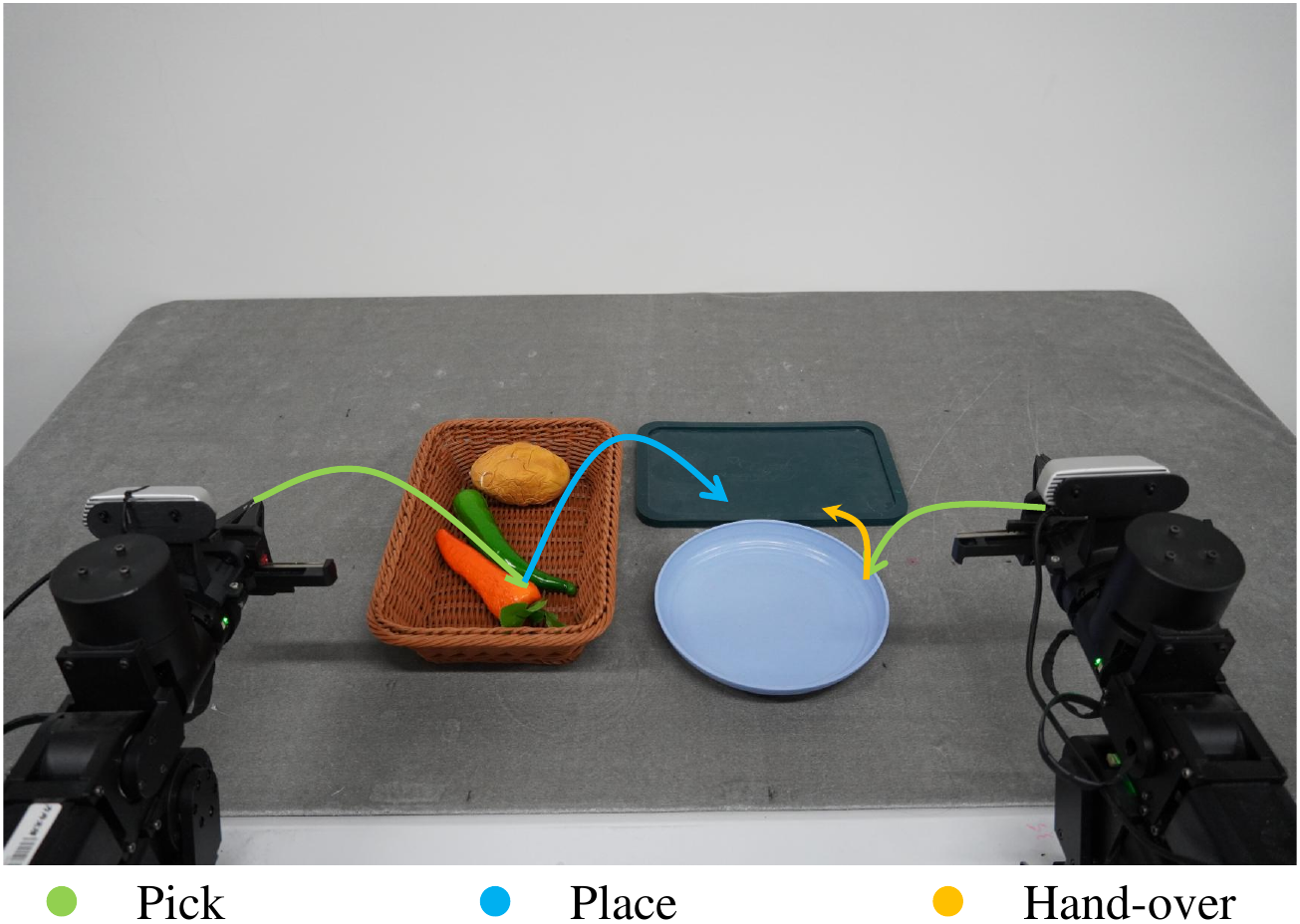
    }
    \label{fig:skill_example}
  }
  \caption{Analysis and visualization of skill distribution across different robotic embodiments.}
  \label{fig:skill_analysis}
  \vspace{-1em}
\end{figure}

\subsection{Quantitative Analysis}

\begin{figure*}[t]
  \centering
  \subfloat{\includegraphics[width=0.97\textwidth]{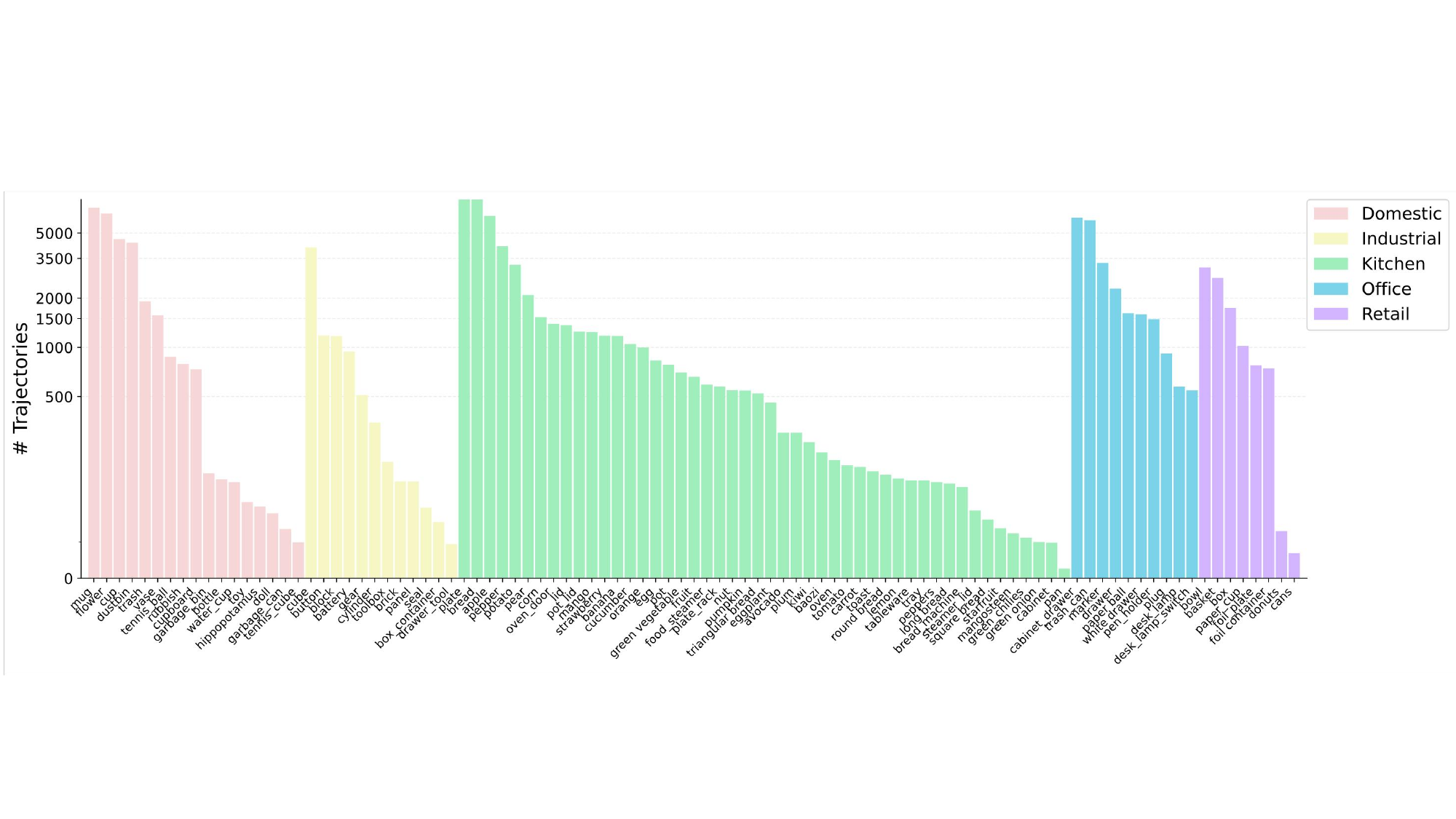}}
  \caption{
    Distribution of objects in \ours, categorized as domestic, industrial, kitchen, office, and retail. 
    The y-axis uses a logarithmic scale for counts above 500, with exact numbers shown for values exceeding it.
  }
  \label{fig:diverse_objects}
  \vspace{-1em}
\end{figure*}

\textbf{Heterogeneous Embodiments.}
A manipulation dataset with different robotic embodiment types improves generalization to various actions and joint DoFs in downstream tasks.
We select four mainstream hardware platforms, each paired with different actuators:
the single-arm robots, Franka Emika Panda and UR5e with grippers;
the dual-arm robot AgileX Cobot Magic V2.0 with grippers;
and \humantk{a humanoid robot}{the humanoid robot \nrobot} equipped with dexterous hands.
Figure~\ref{fig:teaser}(a) shows the distribution of trajectories across different embodiments in our dataset. 
Franka accounts for 49.2\% of the total trajectories, with over 26,070 simulation-based trajectories from our digital twin environment and 26,866 real-world trajectories collected via human teleoperation.
The remaining three embodiments consist solely of real-world demonstrations.
Specifically, the dual-arm data enhances the dataset’s diversity and complexity, supporting the training of coordination skills and more long-horizon tasks.
Additionally, the humanoid robot with dexterous hands, which constitutes 17.8\% of the trajectories, can perform a series of complex, human-like manipulation skills.
The heterogeneous set of embodiment data collected under a unified standard can provide pretraining data for policy models with different action spaces~\citep{liu2024rdt,kim2024openvla}, as well as experimental data for the cross-embodiment transfer research~\citep{xu2023xskill,chen2024mirage}.

\textbf{Tasks with Various Horizon Lengths.}
In addition to the diversity across robot, the varied task horizons in the dataset directly impact the temporal generalization capabilities of policies in real-world scenarios.
We calculate the average task horizon (the number of time steps in one trajectory) for each embodiment, as shown in Figure~\ref{fig:teaser}(b). 
Tasks collected by Franka and UR have shorter trajectories (fewer than 200 time steps), making them ideal for training primitive skills.
In contrast, tasks from \humantk{the humanoid robot}{\nrobot} and AgileX have longer trajectories (over 500 time steps), better suited for long-horizon task training and skill composition.
Since each task involves a varying number of skills, we computed the skill number distribution for each embodiment in Figure~\ref{fig:skill_analysis}(a), offering a clearer view of task horizons.
AgileX tasks typically involve two or more combined skills, while \humantk{humanoid robot}{\nrobot} tasks vary in length, with some incorporating up to five skills per task.
To provide a clearer explanation of long-horizon task construction, we show an AgileX task involving three skills and visualize its dual-arm trajectory in Figure~\ref{fig:skill_analysis}(b).
First, the left and right arms perform the pick skill on the carrot and blue plate, respectively.
Next, the left arm hands the carrot to the right arm's plate.
Finally, the right arm places the blue plate onto the black plate.
The entire process involves complex coordination and long-horizon manipulation.

\textbf{Task Classification.}
Unlike the previous dataset categorizing tasks based on de-duplicated verbs~\citep{khazatsky2024droid}, we categorize tasks by summarizing the manipulation skills from task language descriptions, considering various axes such as actions, objects, and trajectory horizons.
Each trajectory may belong to multiple task types, with only the primary type being counted for each trajectory.
As shown in Figure~\ref{fig:teaser}(c), tasks are categorized into six types:
\begin{enumerate}
  \item \textbf{Articulated Manipulations (Artic. M.):}
    Opening, closing, and turning on or off objects with articulated joints;
  \item \textbf{Coordination Manipulations (Coord. M.):}
    Dual-arm coordination between the robot's arms;
  \item \textbf{Basic Manipulations (Basic M.):}
    Fundamental skills like grasping, holding, lifting, and placing;
  \item \textbf{Multiple Object Interactions (Obj. Int.):}
    Interaction with multiple objects, e.g., pushing one cube across another;
  \item \textbf{Precision Manipulations (Precision M.):}
    Complex manipulation and control skills, such as pouring liquid into a cup or inserting a battery;
  \item \textbf{Scene Understanding (Scene U.):}
    Actions with major challenges related to the semantic understanding of the scene, like closing the upper drawer from the right side or placing four large blocks of different colors into corresponding colored boxes.
\end{enumerate}

By breaking down the language descriptions into fine-grained tasks based on verb-noun combinations, \ours includes {\ntasks} distinct tasks.
In summary, \ours encompasses a range of skills (i.e., verbs from descriptions) beyond basic manipulations, significantly enhancing the policy model's manipulation robustness in handling complex and long-horizon tasks.

\textbf{Diverse Objects.}
A generalized policy needs to learn not only a variety of task skills but also how to execute each skill consistently when interacting with different objects.
\ours includes over {\nobjs} object categories from five usage scenarios, as shown in Figure~\ref{fig:teaser}(d), covering most daily life settings: domestic, industrial, kitchen, office, and retail.
To provide a detailed overview, we summarize trajectories for all objects categorized by usage scenario in Figure~\ref{fig:diverse_objects}. 
In each scene, we design multiple tasks involving a variety of objects.
Specifically, in the kitchen, the dataset includes common food items such as strawberries, eggs, bananas, and pears, along with articulated objects like oven doors and bread machines;
Domestic scenarios feature both rigid objects like tennis balls and deformable objects like toys;
Office and industrial scenarios include small objects that require precise control, such as batteries and gears. 
This wide variety of objects increases the dataset's complexity and supports better generalization to unseen objects in downstream tasks.

\subsection{Qualitative Analysis}

\begin{figure}[t]
  \centering
  \subfloat{\includegraphics[width=0.95\columnwidth]{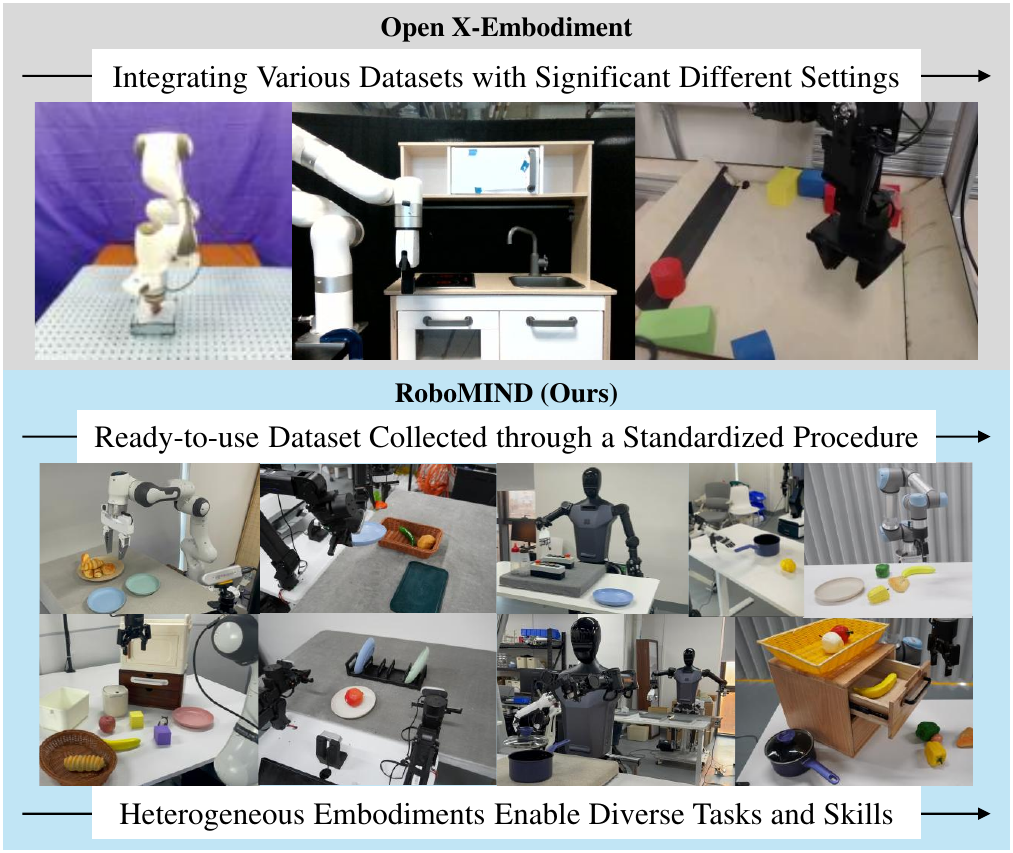}}
  \caption{
    Comparison between Open X-Embodiment and \ours.
    \ours features heterogeneous embodiments with diverse tasks and skills while providing ease of use due to standardized settings.
  }
  \label{fig:dataset_cmp}
  \vspace{-1em}
\end{figure}

\textbf{Standardized Settings.}
\ours features standardized settings to form a large-scale real-world manipulation dataset.
As shown in Figure~\ref{fig:dataset_cmp}, we compare our dataset with Open~X-Embodiment, another large-scale robotic learning dataset.
Although Open~X-Embodiment contains a vast amount of data, the significantly different settings make it difficult to learn efficient manipulation policies across the entire dataset.
In contrast, \ours is collected through a carefully designed standardized procedure, making it ready-to-use for other roboticists.
Meanwhile, its heterogeneous embodiments, diverse tasks, and various skills are suitable for training generalizable policies, whether for primitive skills or long-horizon manipulations.

\begin{figure}[t]
  \centering
  \subfloat{\includegraphics[width=0.48\textwidth]{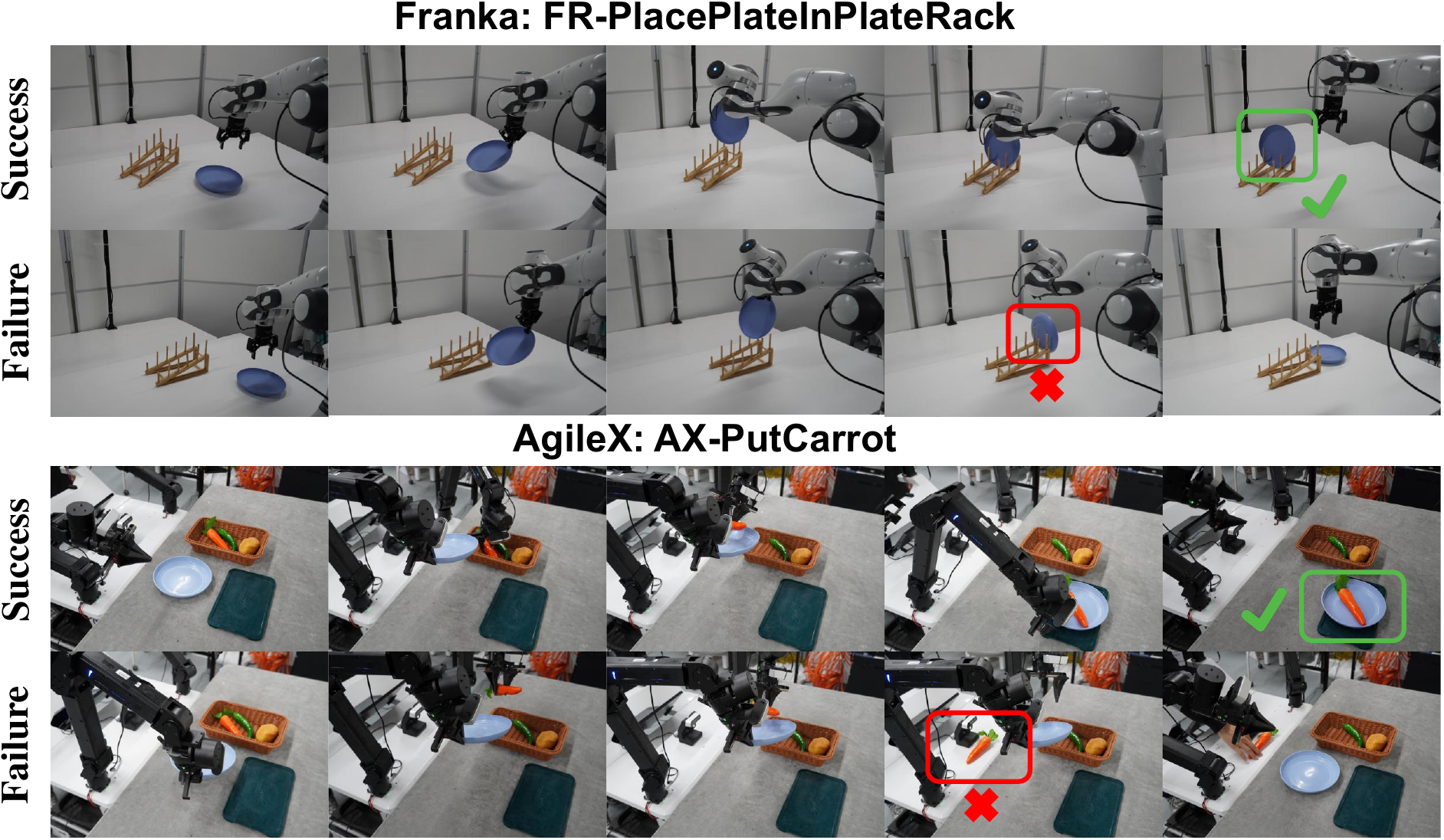}}
  \caption{
    Visualization of failed data collection cases.
    We present two examples of failure from Franka and AgileX.
    In the \texttt{FR-PlacePlateInPlateRack} task (the second row), the Franka arm fails to align with the slot, causing the plate to slip due to operator interference.
    In the \texttt{AX-PutCarrot} task (the fourth row), the AgileX gripper unexpectedly opens, dropping the carrot.
    These failure cases were filtered out during quality inspection to maintain the dataset quality.
  }
  \label{fig:failure_demos}
  \vspace{-1em}
\end{figure}

\textbf{Failure Case Demonstrations.}
We also release 5k trajectories of the robot task failure cases.
The failure cases documented include scenarios where different types of humane operators failed to complete their assigned tasks, as well as instances where robots encountered failures during the execution of operational tasks.
We present the visualization examples from the Franka and AgileX robots of these failure cases in Figure~\ref{fig:failure_demos}.
For the \texttt{FR-PlacePlateInPlateRack} task performed by Franka, a successful execution shows the robotic arm accurately placing a plate into the plate rack.
In the failure case, the arm fails to locate the correct slot position, causing the plate to slip out of the rack, likely due to visual occlusion or interference from the operator.
For the \texttt{AX-PutCarrot} task performed by AgileX, successful execution demonstrates the robot's collaborative manipulation to place a carrot onto the plate. 
In the failure case, the robot's gripper unexpectedly opens, causing the carrot to drop prematurely and resulting in task failure-presumably due to accidental gripper activation by the operator.
During the data quality inspection process, these failed trajectories are identified, categorized, and documented, further enhancing the overall quality of the dataset.

The failure data is intended to advance research in areas like failure detection and recovery, data augmentation, and reward generation for reinforcement learning. 
For instance, training a binary classifier on success/failure data can aid failure detection,
SSDF~\citep{wu2024swbt} filters high-quality data segments from failure data,
and the RLHF approach~\citep{bai2022training} uses failure data as negative examples to learn accurate patterns.
Our failure data can be used seamlessly in these works.

\section{Analyzing Robot Learning with \ours}

\begin{figure}[tb]
  \centering
  \subfloat{\includegraphics[width=1.0\columnwidth]{\humantk{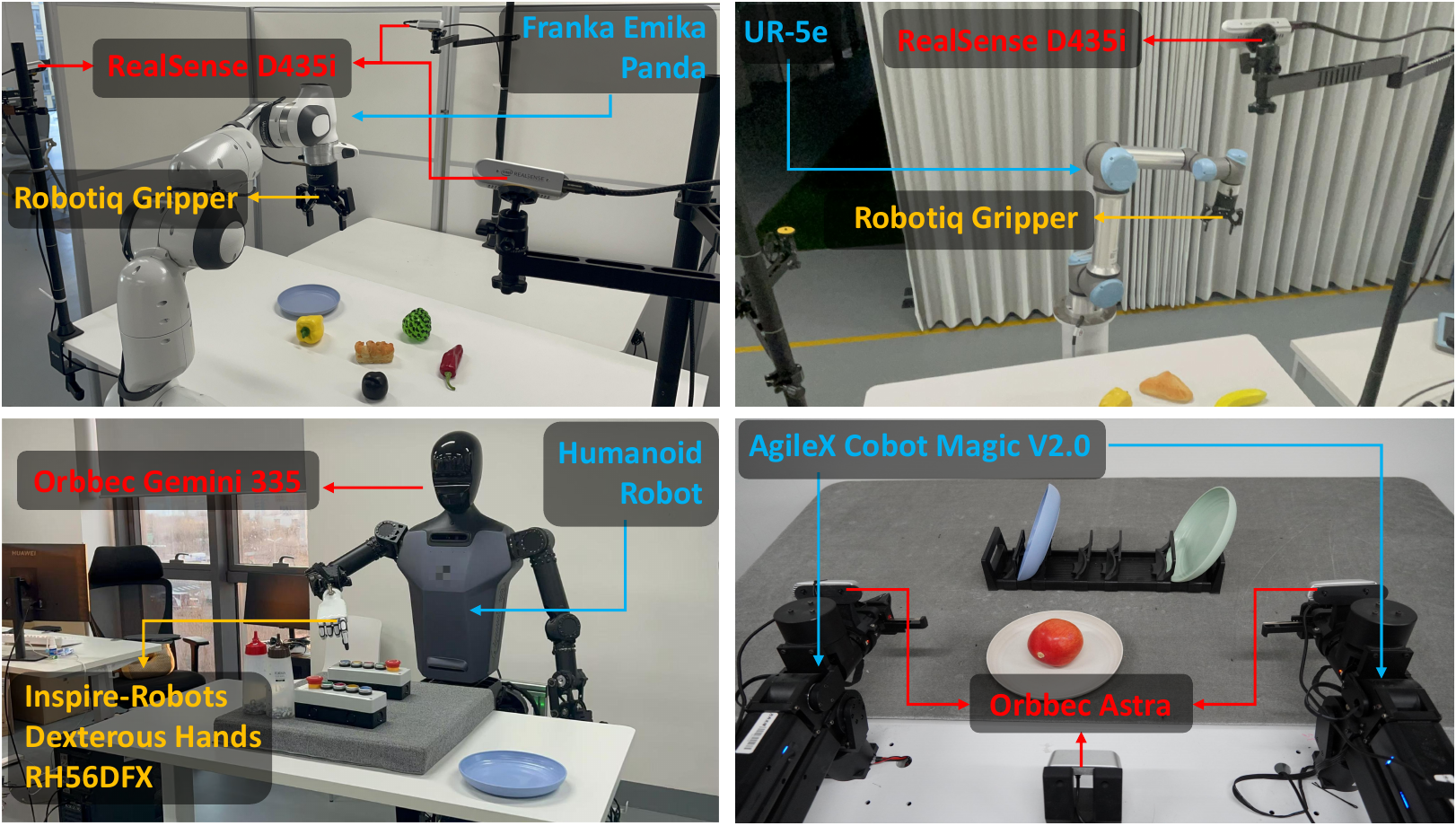}{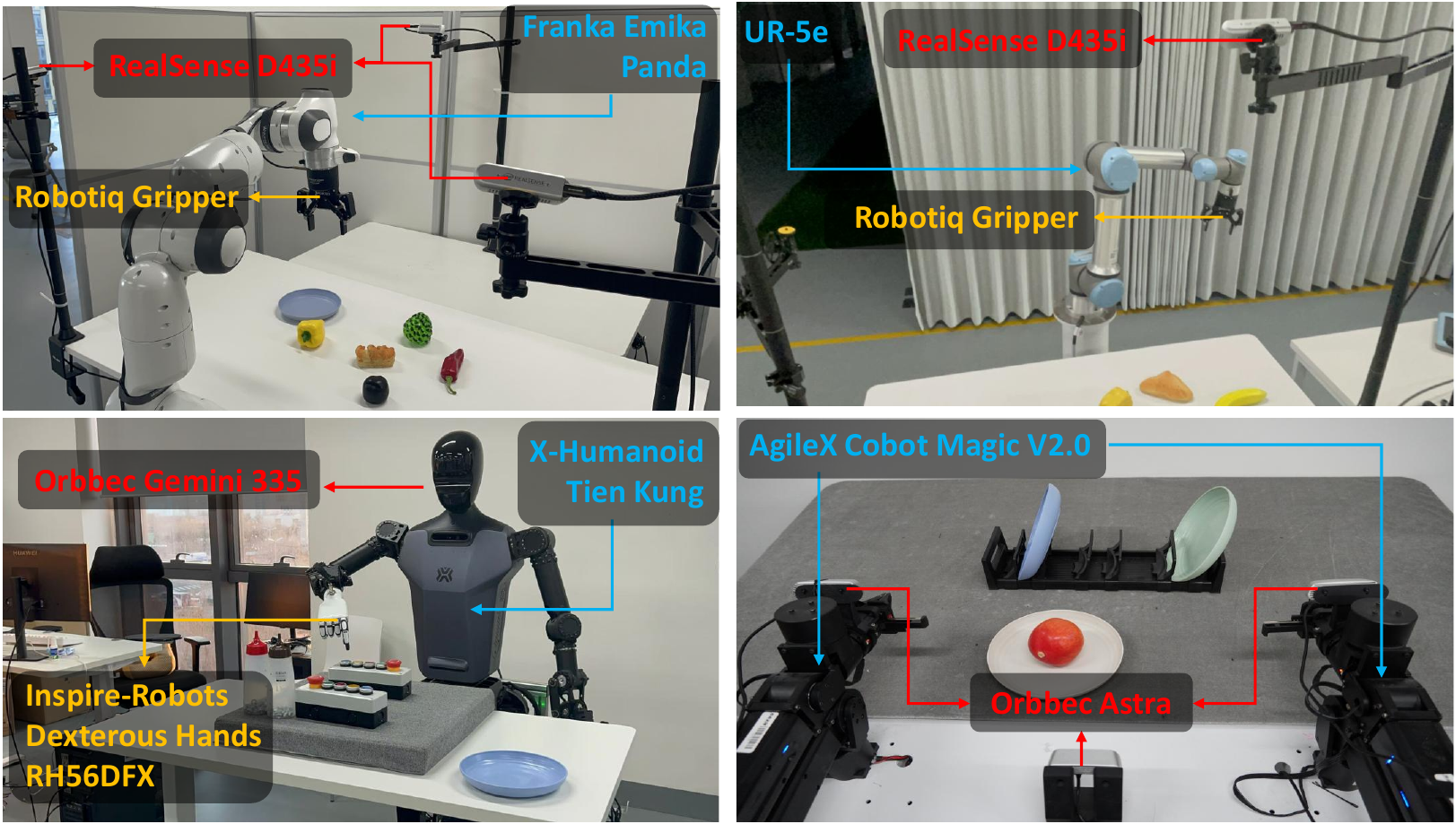}}}
  \caption{
    Robotic real-world setup.
    For the Franka robot, we use cameras positioned at the top, left, and right viewpoints to record the visual information of the task trajectories.
    For the \humantk{humanoid}{\nrobot} and AgileX robots, we use their built-in cameras to record visual information.
    For the UR robot, we use an external top camera.
  }
  \label{fig:hardware_setup}
  \vspace{-1em}
\end{figure}

\begin{figure*}[tb]
  \centering
  \subfloat{\includegraphics[width=0.97\textwidth]{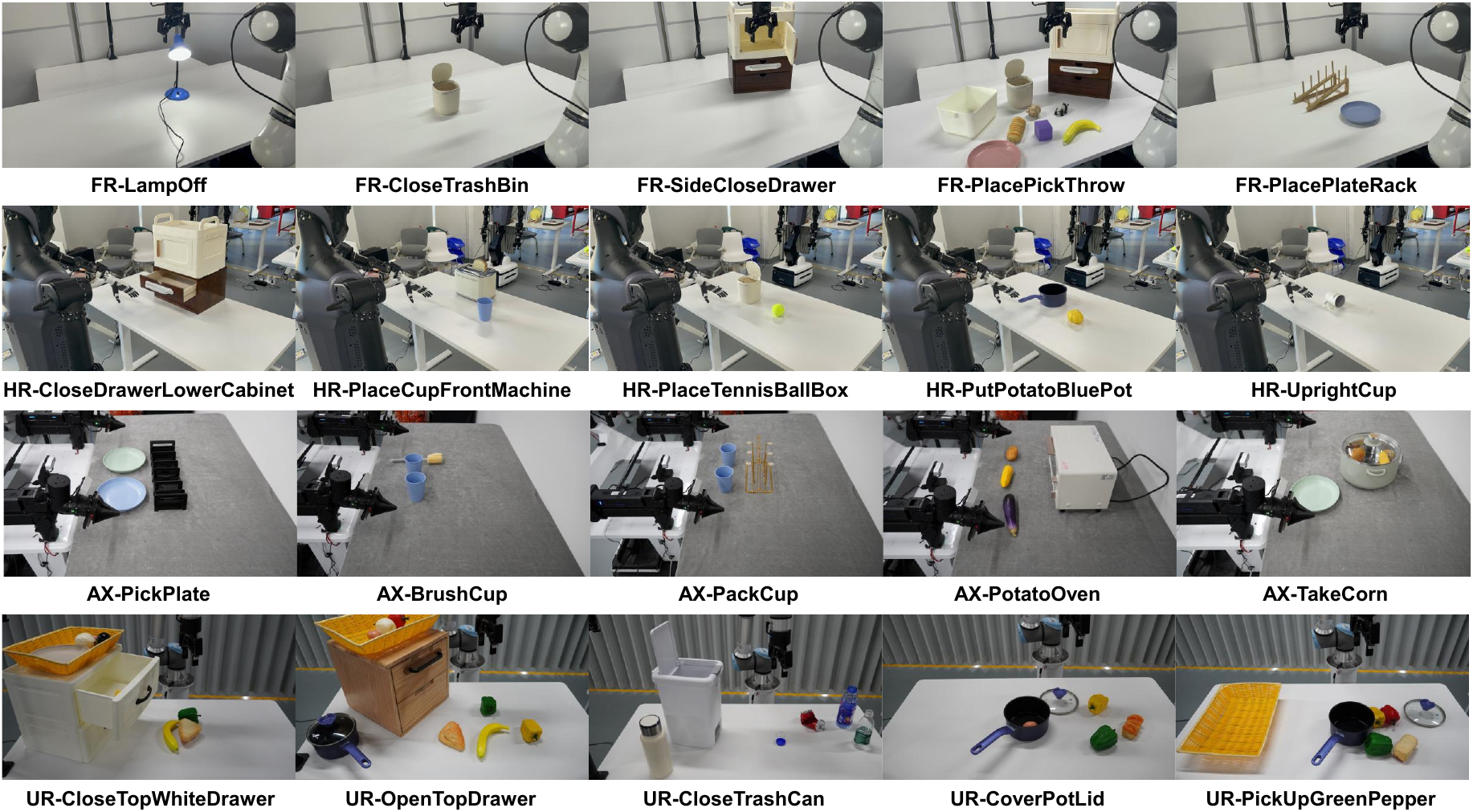}}
  \caption{
    Diverse task examples across 4 robotic embodiments in \ours. 
    The dataset features tasks performed by four distinct robotic embodiments: 
    Franka (the first row), \humantk{the humanoid robot}{\nrobot} (the second row), AgileX (the third row), and UR (the fourth row).
    For each robotic embodiment, we have selected 5 representative task scenarios.
  }
  \label{fig:il_tasks}
  \vspace{-1em}
\end{figure*}

Following the detailed description of \ours's collection process and an in-depth analysis of its characteristics, we conducted a series of comprehensive experiments employing various robot manipulation learning methods.
\ours serves as a benchmark to evaluate the performance and limitations of these methods. 
In the subsequent experiments, we assessed the performance of single-task imitation learning models (ACT~\citep{zhao2023learning}, Diffusion Policy~\citep{chi2023diffusion_policy}, and BAKU~\citep{haldar2024baku}), 
as well as VLA large models (RDT-1B~\citep{liu2024rdt}, OpenVLA~\citep{kim2024openvla}, and CrossFormer~\citep{doshi2024scaling}), which can perform multiple tasks with \ours.
Subsequently, we validated the ability of the VLA models to generalize across various scenarios and manipulate different types of objects. 
Additionally, we applied \ours to pre-train the aforementioned VLA large models, demonstrating that \ours also facilitates cross-embodiment task execution for the VLA large models.
Finally, we provided some failure case analyses and validated the effectiveness of our digital twin simulation data via co-training.

\subsection{Experiment Setup}
\label{subsec:hardware_setup}

\textbf{Real-world Robotic Setup.}
Our real-world robotic setup is shown in Figure~\ref{fig:hardware_setup}. 
The robotic platforms used in this study are equipped as follows:
\textbf{(1) Franka Emika Panda}~\citep{franka_site} features three Intel RealSense D435i cameras~\citep{intel_d435i} (left, top, and right) with resolutions of 480 $\times$ 640, 720 $\times$ 1280, and 480 $\times$ 640 pixels, respectively, and a Robotiq gripper.
\textbf{(2) \humantk{The humanoid robot}{\nrobot}}\humantk{}{~\citep{tien_kung_site}} utilizes two Inspire-Robots RH56DFX dexterous hands and Orbbec Gemini 335 cameras~\citep{orbbec_gemini_335} on the head and chest, both at 480 $\times$ 640 resolution.  
\textbf{(3) AgileX Cobot Magic V2.0}~\citep{agilex_site} is fitted with two hand-eye Orbbec Astra cameras~\citep{orbbec_astra_series} and one front-facing camera, all at 480 $\times$ 640 resolution. 
\textbf{(4) UR5e}~\citep{ur5e_site} is paired with a top-mounted Intel RealSense D435i camera at 480 $\times$ 640 resolution and employs a Robotiq gripper.

\textbf{Representative Tasks.}
\ours encompasses a diverse collection of {\ntasks} distinct manipulation tasks collected across four different robot embodiments. 
Representative examples of these tasks are illustrated in Figure~\ref{fig:il_tasks}.
Below, we provide a representative task for each robot to elucidate the nomenclature and functionality associated with these operations.
\begin{itemize}
  \item \texttt{FR-SideCloseDrawer}. 
    This task requires the Franka robotic arm to locate the outer edge of a cabinet door accurately. 
    The robot needs to make contact with the door edge and push it along a curved path. 
    The goal is to completely close the cabinet door.
  \item \texttt{HR-UprightCup}.
    In this task, \humantk{the}{the \nrobot} humanoid robot needs to grasp a cup that is lying on its side. 
    The robot must then execute a 90-degree rotation movement to bring the cup to an upright position. 
    Finally, it needs to place the upright cup on the table surface gently.
  \item \texttt{AX-TakeCorn}.
    For this task, the AgileX robot must first use its left hand to locate and open the pot lid. 
    The robot then extends its right hand into the pot to grip the corn. 
    Finally, it needs to carefully lift the corn out of the pot and place it onto a plate.
  \item \texttt{UR-CloseTopWhiteDrawer}.
    This task is performed by the UR5e robot, wherein the robot is required to close the uppermost drawer of a set of stacked white drawers.
\end{itemize}

\subsection{Single-task Imitation Learning Models}
\label{Sec:exp-single}

\textbf{Experimental Task Design.}
We carried out our single-task experiments on a large set of single tasks.
We used a total of 45 tasks which were grouped based on the robots performing them. 
Franka, \humantk{the humanoid robot}{\nrobot}, AgileX, and UR5e carried out 15, 10, 15, and 5 tasks respectively. 
We carefully chose these tasks to include a wide variety of actions collected in {\ours}.
These actions ranged from simpler tasks like picking up different objects and placing them in specified spots, to more complex tasks like pulling and pushing articulated objects.
Additional tasks involved dual-arm coordination and precise operations, posing further challenges to the learning capabilities of the models.

\textbf{Training and Evaluation Setup.}
In terms of the imitation learning algorithms, we used three well-known and commonly used methods: 
ACT~\citep{zhao2023learning}, Diffusion Policy~\citep{chi2023diffusion_policy}, and BAKU~\citep{haldar2024baku}. 
For ACT and BAKU, we followed the default model settings as recommended in their original papers.
For Diffusion Policy, we followed the implementation in DROID~\citep{khazatsky2024droid}.
Using the three algorithms, we trained the single-task model from scratch for each dataset. 
After training, we directly deployed the models in real-world environments for evaluation.
We assessed the performance of each model using its success rate in the tasks. 
Each model was tested ten times, and the testers recorded the success or failure of each test and the reasons if there were any failures. 
This thorough process gave us valuable insights for further developments.

\begin{figure*}[tb]
  \centering
  \subfloat{
    \includegraphics[width=0.98\textwidth]{
      \humantk{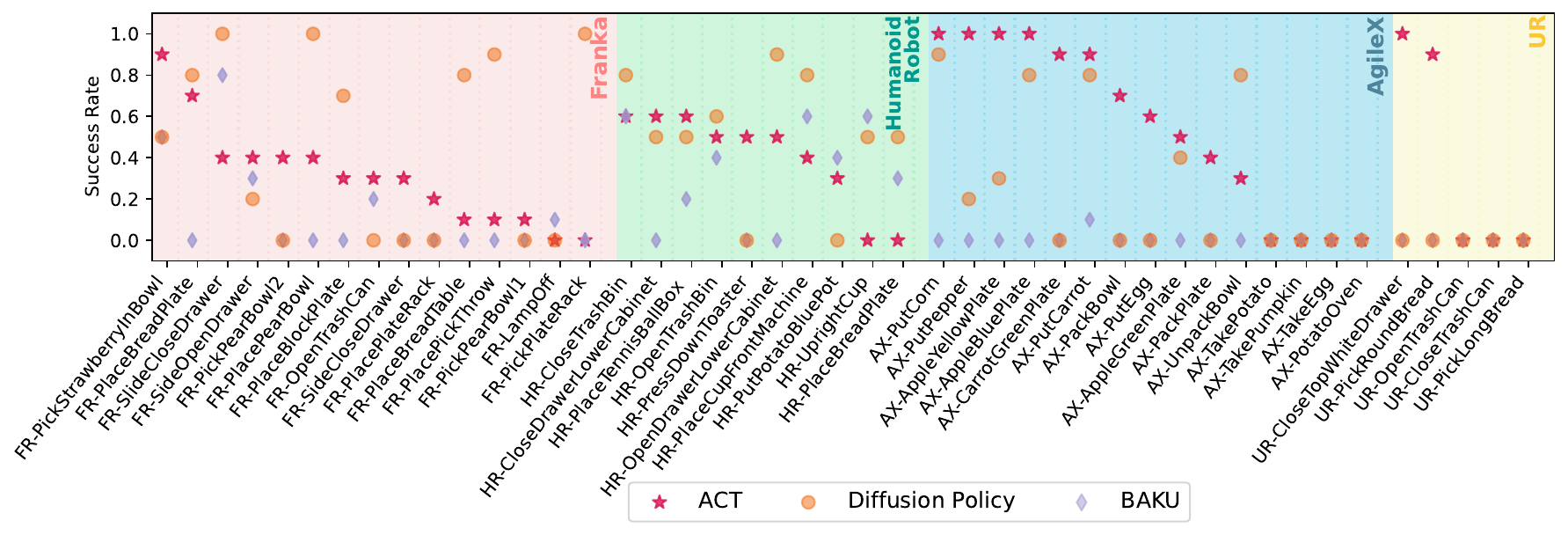}{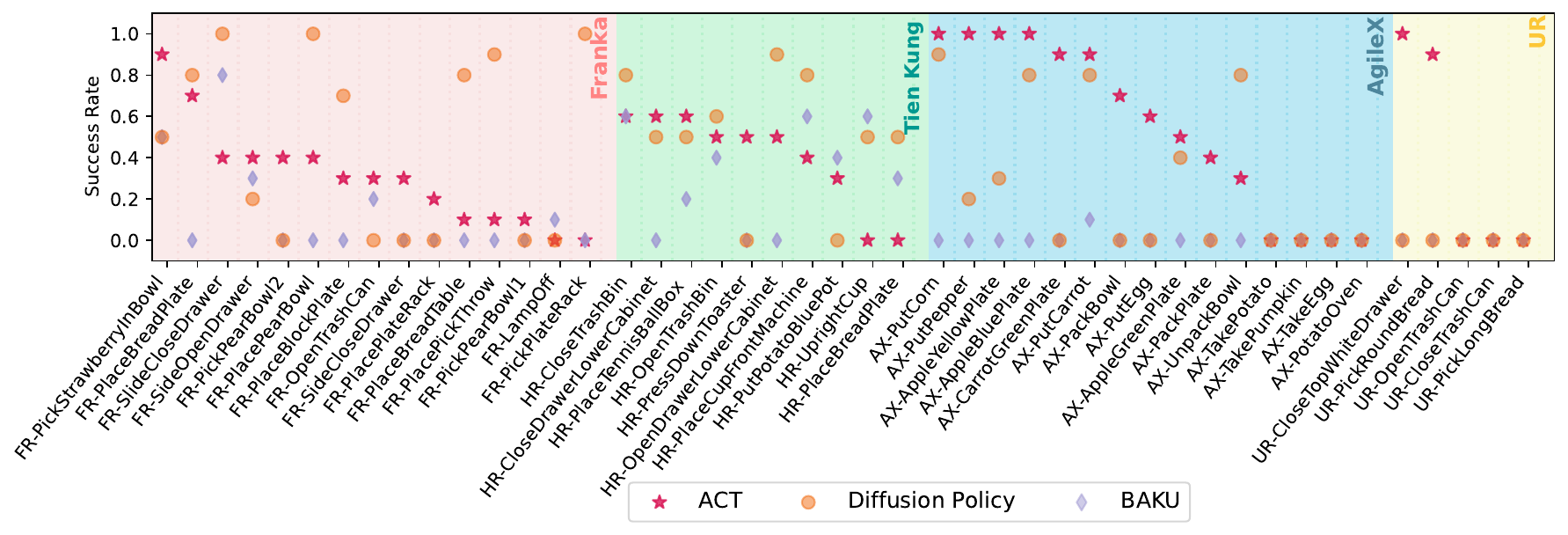}
    }
  }
  \caption{
    Success rates of ACT, Diffusion Policy, and BAKU on \ours.
  }
  \label{fig:il_results}
  \vspace{-1em}
\end{figure*}

\begin{figure*}[tb]
  \centering
  \subfloat{\includegraphics[width=0.98\textwidth]{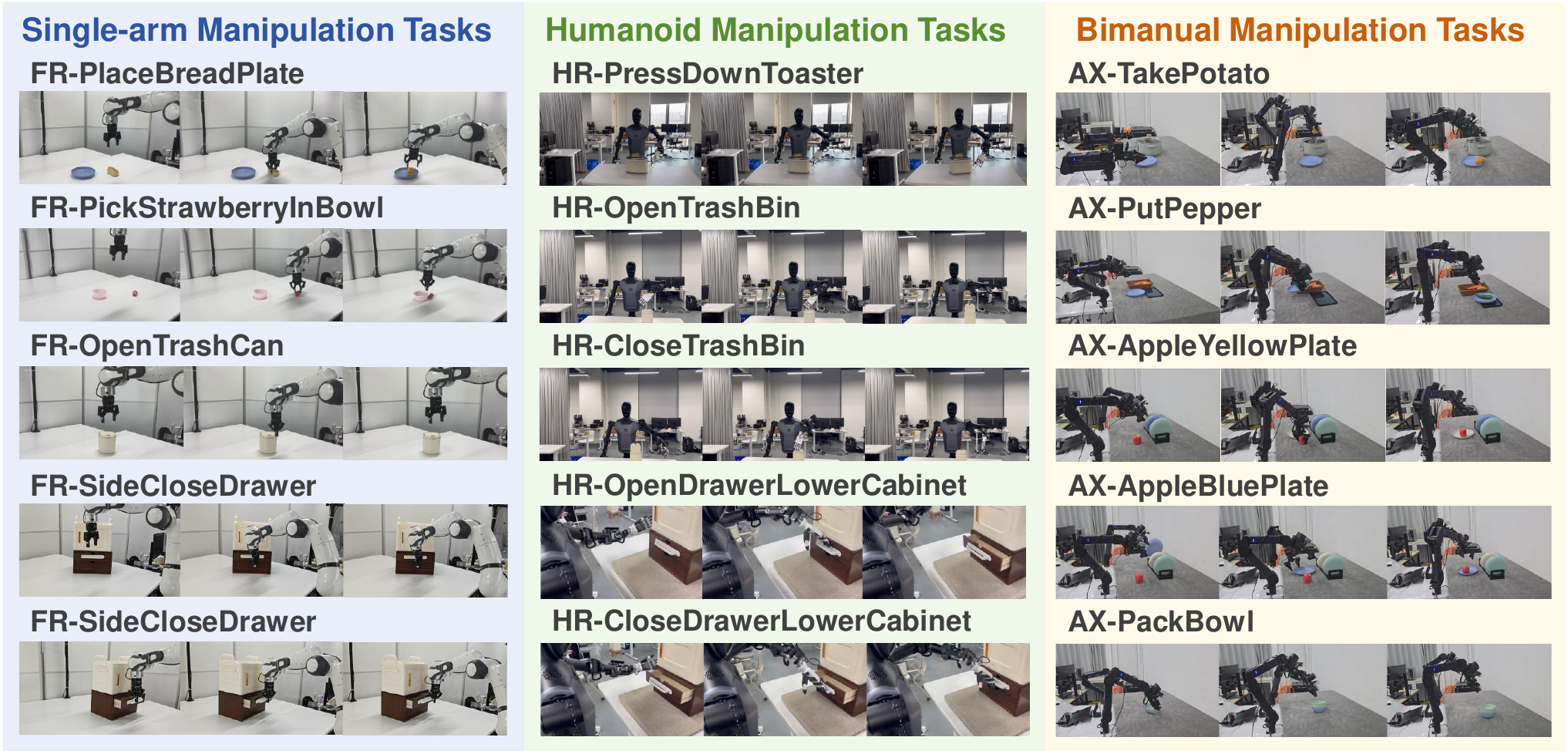}}
  \caption{
    Visualization of the selected tasks on single-arm, dual-arm, and humanoid robots used in experiments of the vision-language-action models.
  }
  \label{fig:all_robot_and_task}
  \vspace{-1em}
\end{figure*}

\textbf{Experimental Results.}
Figure~\ref{fig:il_results} presents the performance of ACT~\citep{zhao2023learning}, Diffusion Policy~\citep{chi2023diffusion_policy}, and BAKU~\citep{haldar2024baku} across 45 tasks using four types of robots, evaluated in terms of the success rate. 
In Figure~\ref{fig:il_results}, we found that ACT achieves an average success rate of 55.3\% across 15 tasks on AgileX, outperforming Franka (30.7\%), UR5e (38.0\%), and \humantk{the humanoid robot}{\nrobot} (34.0\%).  
Additionally, ACT also showed promising results on several humanoid robot tasks, including a 60\% success rate on \texttt{HR-CloseDrawerLowerCabinet}. 
These results not only illustrated that ACT shows robust performance in complex dexterous hand manipulation tasks but also underscored the high quality of data gathered in {\ours}.
Similarly, Diffusion Policy also demonstrated its capacity to learn complex tasks, outperforming ACT in several tasks on Franka and \humantk{the humanoid robot}{\nrobot}.
Therefore, we believe that the single-arm, dual-arm, and dexterous hand datasets in \ours can serve as high-quality training sets to improve the performance of single-task imitation learning, thereby advancing the development of the entire imitation learning field.
On the other hand, BAKU exhibits lower success rates across most tasks. 
This discrepancy could be attributed to the hyper-parameter settings from the original BAKU paper, which is primarily optimized for simulation environments rather than real-world robotic platforms tested in our experiments. 
The significant performance gap underscores the challenges in directly transferring models from simulated settings to physical robots.

\subsection{Vision-Language-Action Large Models}
\label{EXP:MUL}

\textbf{Experimental Task Design.}
This section seeks to examine the performance of VLA large-parameter robot model when applied to \ours. 
We picked fifteen tasks performed by different types of robots from the single-task imitation learning experiments.
Figure~\ref{fig:all_robot_and_task} illustrates the tasks we chose for Franka single-arm robot, \humantk{the}{the \nrobot} humanoid robot, and the AgileX dual-arm robot.
For \textbf{the Franka single-arm robot}, these selected tasks encompass common robotic arm operations, such as picking and placing, pushing and pulling, along with more nuanced tasks that require precise manipulation, including picking objects of varying sizes and accurately positioning the robotic arm to open a trash bin lid.
For \textbf{\humantk{the}{the \nrobot} humanoid robot}, the tasks are divided into two main categories.
The first category consists of tasks similar to those performed by the single-arm Franka robot, which are intended to evaluate the model's performance across different robot types. 
The second category involves using the humanoid robot's dexterous hands to perform precise operations, such as flipping a toaster switch to toast bread, to assess the model's accuracy in positioning and manipulation. 
For \textbf{the AgileX dual-arm robot}, we chose dual-arm tasks that involve coordinated actions, such as the left arm retrieving a plate from a rack and the right arm placing an apple on the plate. 
This selection emphasizes the unique capabilities and coordination required in dual-arm operations.

\textbf{Training and Evaluation Setup.}
We evaluated the performance of three models (OpenVLA~\citep{kim2024openvla}, RDT-1B~\citep{liu2024rdt}, and CrossFormer~\citep{doshi2024scaling}) fine-tuned by the demonstrations from \ours in completing various real-world tasks.
Given that the VLA large model exhibits excellent generalization performance, we employed an aggregated dataset sourced from multitask demonstrations for fine-tuning the VLA models. 
Specifically, we took the official pre-trained VLA models and fine-tuned them on the multitask datasets for each type of robot, and evaluated their performance on each individual task to determine the extent of generalization achieved, by conducting ten trials for each task.
We tested ten trials for each experiment.
For OpenVLA~\citep{kim2024openvla}, which involves fine-tuning the Llama~2 model~\citep{touvron2023llama} using a large robotic dataset and adapting it to be a 7-DoF VLA model, we only tested it on the Franka single-arm robot, since the output of OpenVLA is the condition of one end effector and only supports single-arm manipulations.

\begin{table}[t]
  \centering
  \caption{
    Success rates of the VLA models in the fine-tuning settings using \ours.
    \colorbox[HTML]{E0F4FF}{\textbf{Color boxes}} represent the first best performance in all tables of this paper.
  }
  \resizebox{0.98\columnwidth}{!}{
    \setlength{\tabcolsep}{5pt}
    \begin{tabular}{l|ccc}
      \toprule
      \textbf{Single-arm Manipulation Task} & OpenVLA~\citep{kim2024openvla} & RDT-1B~\citep{liu2024rdt} & CrossFormer~\citep{doshi2024scaling} \\
      \cmidrule(lr){1-4}
      \texttt{FR-PlaceBreadPlate}      & 4/10             & \bbluecell{7/10} & 0/10             \\
      \texttt{FR-PickStrawberryInBowl} & 0/10             & \bbluecell{4/10} & 0/10             \\
      \texttt{FR-OpenTrashCan}         & \bbluecell{3/10} & \bbluecell{3/10} & 0/10             \\
      \texttt{FR-SideCloseDrawer}      & \bbluecell{7/10} & 6/10             & 2/10             \\
      \texttt{FR-SideOpenDrawer}       & 0/10             & 2/10             & \bbluecell{4/10} \\
      \midrule
      \textbf{Humanoid Manipulation Tasks}  & OpenVLA~\citep{kim2024openvla} & RDT-1B~\citep{liu2024rdt} & CrossFormer~\citep{doshi2024scaling} \\
      \cmidrule(lr){1-4}
      \texttt{HR-OpenDrawerLowerCabinet}  & -- & \bbluecell{5/10} & \bbluecell{5/10} \\
      \texttt{HR-CloseDrawerLowerCabinet} & -- & \bbluecell{5/10} & 3/10             \\
      \texttt{HR-OpenTrashBin}            & -- & \bbluecell{2/10} & 0/10             \\
      \texttt{HR-CloseTrashBin}           & -- & \bbluecell{3/10} & \bbluecell{3/10} \\
      \texttt{HR-PressDownToaster}        & -- & 3/10             & \bbluecell{7/10} \\
      \midrule
      \textbf{Bimanual Manipulation Task}   & OpenVLA~\citep{kim2024openvla} & RDT-1B~\citep{liu2024rdt} & CrossFormer~\citep{doshi2024scaling} \\
      \cmidrule(lr){1-4}
      \texttt{AX-TakePotato}       & -- & \bbluecell{6/10}  & 0/10 \\
      \texttt{AX-PutPepper}        & -- & \bbluecell{9/10}  & 0/10 \\
      \texttt{AX-AppleYellowPlate} & -- & \bbluecell{10/10} & 0/10 \\
      \texttt{AX-AppleBluePlate}   & -- & \bbluecell{6/10}  & 0/10 \\
      \texttt{AX-PackBowl}         & -- & \bbluecell{8/10}  & 0/10 \\

      \bottomrule
    \end{tabular}
  }
  \label{tab:vla}
  \vspace{-1em}
\end{table}

\textbf{Experimental Results.}
Table~\ref{tab:vla} presents the success rates for various robot tasks performed using the three different VLA models.
The experimental results show that the VLA large models fine-tuned on expert demonstrations from \ours performed well across various different robot tasks.
The fine-tuned RDT-1B, compared to CrossFormer and OpenVLA, demonstrated significantly enhanced performance in executing tasks across a range of robot models.
This improvement is especially notable for dual-arm manipulation tasks, where RDT-1B excelled.
Although the performance of OpenVLA being inferior to that of RDT-1B, it nonetheless achieved a comparable task success rate for straightforward tasks like \texttt{FR-PlaceBreadPlate} and \texttt{FR-SlideCloseDrawer}.
CrossFormer, after being fine-tuned with \ours, demonstrated performance improvements in tasks executed by single-arm and humanoid robots, compared with no success in all tasks without fine-tuning. 

\begin{table*}[t]
  \centering
  \caption{
    Success rates of the VLA models before and after training with \ours. 
    The notation `(origin)' indicates models fine-tuned directly on the expert multitask dataset without training on \ours, 
    while `(\ours)' denotes models first trained on the entire \ours dataset and subsequently fine-tuned on the expert multitask dataset.
  } 
  \begin{tabular}{l|cccc}
    \toprule
    \textbf{Single-arm Manipulation Task} & RDT-1B (origin) & RDT-1B (\ours) & CrossFormer (origin) & CrossFormer (\ours) \\
    \cmidrule(lr){1-5}
    \texttt{FR-PlaceBreadPlate}      & 7/10 & 9/10             & 0/10 & \bbluecell{10/10} \\
    \texttt{FR-PickStrawberryInBowl} & 4/10 & 6/10             & 0/10 & \bbluecell{8/10}  \\
    \texttt{FR-OpenTrashCan}         & 3/10 & \bbluecell{6/10} & 0/10 & 0/10              \\
    \texttt{FR-SideCloseDrawer}      & 6/10 & \bbluecell{8/10} & 2/10 & \bbluecell{8/10}  \\
    \texttt{FR-SideOpenDrawer}       & 2/10 & \bbluecell{5/10} & 4/10 & 3/10              \\
    \midrule
    \textbf{Humanoid Manipulation Tasks} & RDT-1B (origin)  & RDT-1B (\ours) & CrossFormer (origin) & CrossFormer (\ours) \\
    \cmidrule(lr){1-5}
    \texttt{HR-OpenDrawerLowerCabinet}  & 5/10 & \bbluecell{6/10} & 5/10 & 4/10              \\
    \texttt{HR-CloseDrawerLowerCabinet} & 5/10 & \bbluecell{7/10} & 3/10 & \bbluecell{7/10}  \\
    \texttt{HR-OpenTrashBin}            & 2/10 & \bbluecell{4/10} & 0/10 & \bbluecell{4/10}  \\
    \texttt{HR-CloseTrashBin}           & 3/10 & \bbluecell{4/10} & 3/10 & 3/10              \\
    \texttt{HR-PressDownToaster}        & 3/10 & 4/10             & 7/10 & \bbluecell{10/10} \\
    \midrule
    \textbf{Bimanual Manipulation Task}   & RDT-1B (origin) & RDT-1B (\ours) & CrossFormer (origin) & CrossFormer (\ours) \\
    \cmidrule(lr){1-5}
    \texttt{AX-TakePotato}       & 6/10  & 8/10              & 0/10 & \bbluecell{10/10} \\
    \texttt{AX-PutPepper}        & 9/10  & \bbluecell{10/10} & 0/10 & \bbluecell{10/10} \\
    \texttt{AX-AppleYellowPlate} & 10/10 & \bbluecell{10/10} & 0/10 & 5/10              \\
    \texttt{AX-AppleBluePlate}   & 6/10  & \bbluecell{9/10}  & 0/10 & \bbluecell{9/10}  \\
    \texttt{AX-PackBowl}         & 8/10  & \bbluecell{10/10} & 0/10 & 4/10              \\
    \bottomrule
  \end{tabular}
  \label{tab:robomind}
  \vspace{-1em}
\end{table*}

\subsection{Leveraging \ours to Enhance VLA Large Models}

\textbf{Training and Evaluation Setup.}
Currently, most VLA large models are trained with datasets from robots with arms and grippers and can only be applied to the same types of robots.
It is noting that \ours contains valuable data from diverse robots including \humantk{humanoid robots}{the {\nrobot} humanoid robots} with dexterous hands,
and we applied this dataset in the pre-training of the RDT-1B and CrossFormer models to enhance their ability to handle real-world dexterous manipulation tasks.
After that, we fine-tuned the VLA large models using the expert multitask datasets, which is a small subset (about 1\%) of \ours.
We conducted ten tests for each model on each task, and we compared the results with those from the official pre-trained and expert data fine-tined models.

\textbf{Experimental Results.}
Table~\ref{tab:robomind} presents the experimental results of RDT-1B and CrossFormer that were first trained on the entire \ours dataset and then fine-tuned on the expert multitask dataset, compared to those fine-tuned directly on the expert multitask dataset.
The results show that training different VLA models using the full \ours dataset led to significant improvements in task success rate across a variety of robot tasks.
Especially for dual-arm tasks on the CrossFormer, training with the full \ours dataset significantly enhanced its performance.
The training effect improved from being unable to complete each dual-arm task to achieving nearly every test success on \texttt{AX-TakePotato}, \texttt{AX-PutPepper}, and \texttt{AX-AppleBluePlate}.
For \texttt{HR-PressDownToaster} from humanoid manipulation tasks, training CrossFormer using \ours also achieved a 100\% task success rate.
This improvement underscores the robustness and versatility of \ours in facilitating more effective and reliable robotic manipulations.

\begin{table}[tp]
  \centering
  \caption{
    Success rates of RDT-1B pre-trained with and without the humanoid data.
  }
  \resizebox{0.98\columnwidth}{!}{
    \begin{tabular}{l|cc}
      \toprule
      \textbf{Success Rate}            & \textbf{RDT-1B (w/o Humanoid)} & \textbf{RDT-1B (w/ Humanoid)} \\
      \midrule
      \texttt{FR-PlaceBreadPlate}      & 9/10                           & 9/10                          \\
      \texttt{FR-PickStrawberryInBowl} & 5/10                           & 6/10 (1/10 $\uparrow$)        \\
      \texttt{FR-OpenTrashCan}         & 5/10                           & 6/10 (1/10 $\uparrow$)        \\
      \texttt{FR-SideCloseDrawer}      & 8/10                           & 8/10                          \\
      \texttt{FR-SideOpenDrawer}       & 3/10                           & 5/10 (2/10 $\uparrow$)        \\
      \bottomrule
    \end{tabular}
  }
  \label{tab:humanoid}
  \vspace{-1em}
\end{table}

\textbf{Ablation Studies on Humanoid Data.}
The standardized data across diverse combinations of robotic platforms eases investigations on cross-embodiment generalization and improvements.
For example, it is worth exploring whether the humanoid data is helpful for policy learning for other robots, such as the single-arm Franka.
We conducted an ablation study by pre-training the RDT-1B model using an incomplete \ours dataset excluding the humanoids data (19k of 107k in \ours)  and fine-tuning the model on the expert multitask datasets.
Table~\ref{tab:humanoid} shows that the humanoid data enhances the performance of RDT-1B on the unimanual tasks, especially the difficult tasks.
Overall, training with the full \ours dataset obtained a 13.3\% relative improvement (0.68 v.s. 0.6) of success rates against the incomplete dataset without humanoid data.

\begin{figure*}[t]
  \centering
  \subfloat{\includegraphics[width=0.95\textwidth]{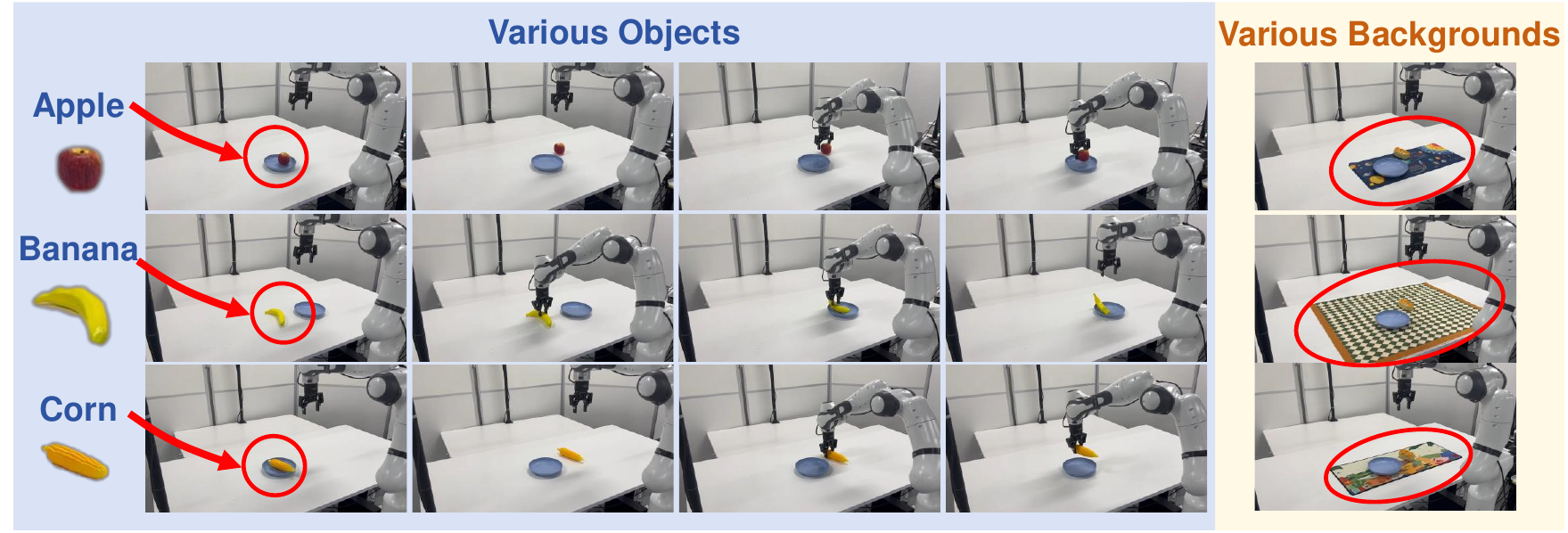}}
  \caption{
    Unseen objects and backgrounds used to evaluate the generalization ability of the VLA large models.
  }
  \label{fig:generalize}
  \vspace{-1em}
\end{figure*}

\subsection{Generalization of VLA Large Models}

\textbf{Evaluation Setup.}
We conducted tests to validate the generalization of using \ours to fine-tune the VLA large models, assessing their ability to generalize across real task scenarios with varying backgrounds and different objects of manipulation.
Specifically, we evaluated the generalization performance on the \texttt{FR-PlaceBreadPlate} task of OpenVLA~\citep{kim2024openvla}, RDT-1B~\citep{liu2024rdt}, and CrossFormer~\citep{doshi2024scaling} funetuned on the Franka multitask dataset in Section~\ref{EXP:MUL}.
Both RDT-1B and CrossFormer are trained on the entire \ours dataset and subsequently fine-tuned using the Franka expert multitask dataset. OpenVLA, in contrast, is directly fine-tuned using the Franka expert multitask dataset.
As shown in Figure~\ref{fig:generalize}, we executed the \texttt{FR-PlaceBreadPlate} task on three tablecloths with different unseen background patterns and replaced the grasped bread object with an apple, a banana, and a corn.
We tested ten trials for each experiment.

\begin{table}[tp]
  \centering
  \caption{
    Generalization results of VLA large models on the \texttt{FR-PlaceBreadPlate}-related tasks.  
  }
  \resizebox{0.98\columnwidth}{!}{
    \begin{tabular}{l|ccc}
      \toprule
      \textbf{Generalization of Backgrounds and Objects} & OpenVLA & RDT-1B           & CrossFormer       \\
      \midrule
      \texttt{FR-PlaceBreadPlate}                        & 4/10    & 9/10             & \bbluecell{10/10} \\
      \cmidrule(lr){1-4}
      \texttt{FR-PlaceCornPlate}                         & 1/10    & 5/10             & \bbluecell{6/10}  \\
      \texttt{FR-PlaceBananaPlate}                       & 1/10    & 6/10             & \bbluecell{9/10}  \\
      \texttt{FR-PlaceApplePlate}                        & 0/10    & \bbluecell{3/10} & 2/10              \\
      \cmidrule(lr){1-4}
      \texttt{FR-PlaceBreadPlate (Unseen Background 1)}  & 0/10    & 1/10             & \bbluecell{2/10}  \\
      \texttt{FR-PlaceBreadPlate (Unseen Background 2)}  & 0/10    & \bbluecell{1/10} & 0/10              \\
      \texttt{FR-PlaceBreadPlate (Unseen Background 3)}  & 0/10    & \bbluecell{0/10} & 0/10              \\
      \bottomrule
    \end{tabular}
    }
  \label{tab:ge}
  \vspace{-1em}
\end{table}

\textbf{Experimental Results.}
As presented in Table~\ref{tab:ge}, both RDT-1B and CrossFormer exhibited good generalizations for manipulating objects, especially for objects like bananas that are similar in shape to the bread-like objects in the training data.
However, when it comes to generalizing across unseen backgrounds, RDT-1B, OpenVLA, and CrossFormer performed relatively poorly in the \texttt{FR-PlaceBreadPlate} task.

\label{sec:fail}
 \begin{figure}[t]
  \centering
  \subfloat{\includegraphics[width=0.95\columnwidth]{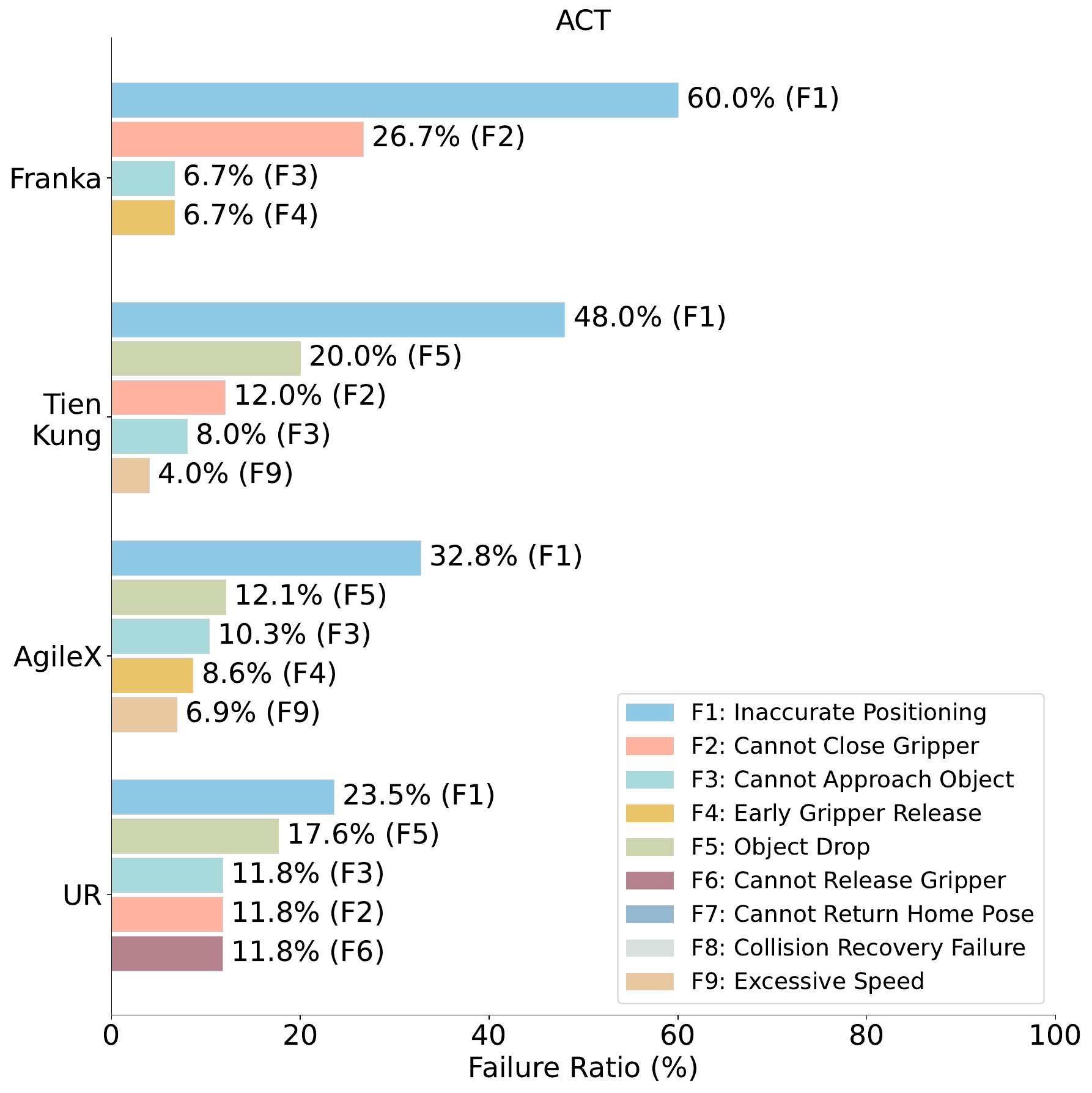}}
  \caption{
    Top five failure reasons for each embodiment of the ACT algorithm.
    The x-axis denotes the proportion for each failure among all unsuccessful test cases.
    The y-axis denotes different embodiments.
  }
  \label{fig:failure_analysis_act_dp}
  \vspace{-1em}
\end{figure}

\subsection{Failure Case Analysis on Real-world Experiments}

During the testing phase, we recorded not only whether the model's task execution was successful but also the reasons for any failures. 
We predefined nine failure categories: 
(1) Inaccurate Positioning; 
(2) Cannot Close Gripper; 
(3) Cannot Approach Object; 
(4) Early Gripper Release; 
(5) Object Drop; 
(6) Cannot Release Gripper; 
(7) Cannot Return to Home Pose; 
(8) Collision Recovery Failure;
(9) Excessive Speed.

In Figure~\ref{fig:failure_analysis_act_dp}, we showed the distribution of failure reasons for the ACT across 45 single tasks performed on the four robotic embodiments, as described in Section~\label{Sec:exp-single}.
We presented the top five most frequent failure reasons for each robotic embodiment.
Firstly, we observe that, for ACT, ``Inaccurate Positioning'' is the most common failure reason across all rollouts. 
For instance, in the humanoid robot tasks, 
failures due to ``Inaccurate Positioning'' accounted for as much as 48\%. 
This highlights the critical importance of accurately positioning the robotic arm in 3D space to execute skills successfully, representing the first step toward achieving task success.
It can be noted that the improper gripper actions, such as ``Cannot Close Gripper'' and ``Object Drop'', were significant contributors to overall task failures. 
This issue arises because the number of frames used for gripper actions is typically limited, thereby complicating the learning process.

From a data perspective, which is often overlooked by researchers and developers, the reasons for failure provide insights into improving data quality.
The collected data frequently fall short of the task designer's expectations due to various factors such as hardware limitations, physical state, external interference, and communication issues.
For instance, inaccurate localization may stem from non-random placement of objects in the dataset, despite instructions for random placement. 
To address this, we can collect additional data from previously neglected locations to better represent the task environment and improve the success rate.
Similarly, gripper non-closure is likely due to the data collector moving too quickly when closing the jaws, resulting in insufficient frames being captured. 
This makes training more challenging.
To mitigate this, we can instruct collectors to slow down during jaw closure to ensure adequate data capture.
By refining data collection practices, we can enhance the robustness and reliability of the imitation learning algorithms, ultimately leading to better performance in real-world applications.

\subsection{Real and Simulation Data}
\label{EXP:Real_Sim_cotrain}

To validate the effectiveness of simulation data in \ours, we conducted two sets of experiments.

\begin{figure}[t]
  \centering
  \subfloat{\includegraphics[width=1.0\columnwidth]{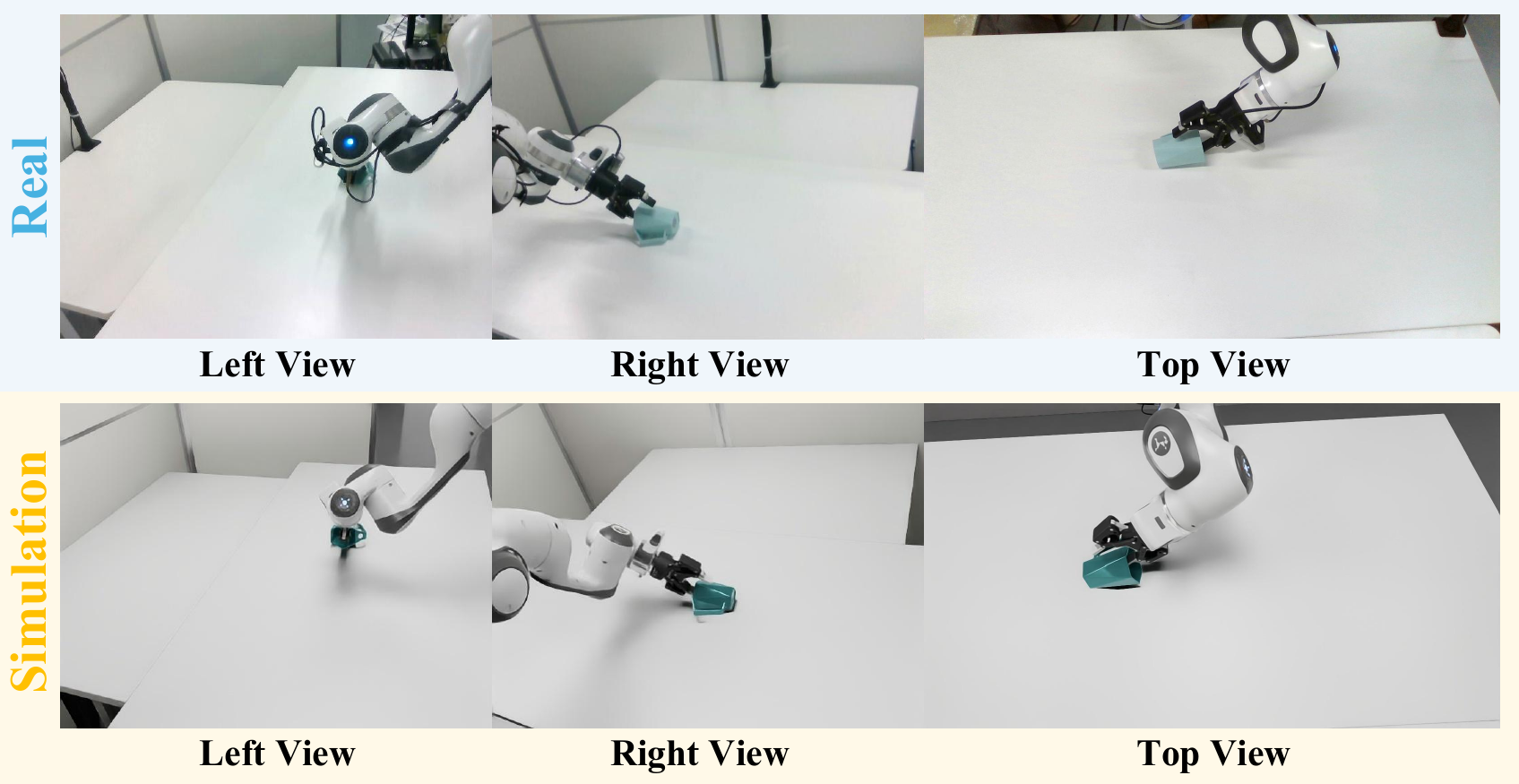}}
  \caption{
    Experimental setup in real-world and simulation environments.
    The top and bottom rows show observations from the left view, right view, and top view in the real-world and simulation environments, respectively.
    We can see that the two environments are very similar, as the simulation environment was constructed to mirror the real environment.
  }
  \label{fig:real_sim_setup}
  \vspace{-1em}
\end{figure}

\begin{figure}[t]
  \centering
  \subfloat{\includegraphics[width=0.97\columnwidth]{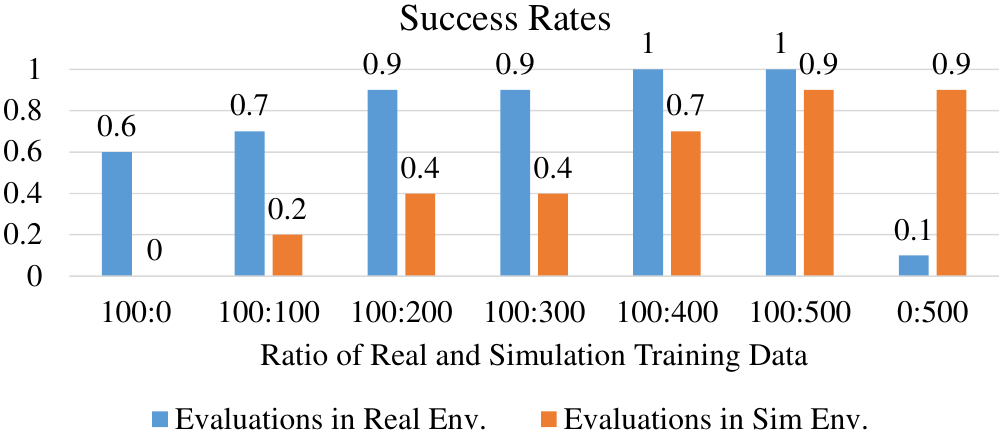}}
  \caption{
    Success rates of models trained with different ratios of real-world and simulation data.
  }
  \label{fig:real_sim_results}
  \vspace{-1em}
\end{figure}

\textbf{Co-training with Real and Simulation Data.}
Firstly, we combined both real-world and simulation data for training.
We selected a complex Franka robotic arm task, \texttt{FR-UprightBlueCup}, which requires the robotic arm to rotate nearly 90 degrees, insert its gripper horizontally into the cup's opening, and restore an overturned cup to its upright position.
As shown in Figure~\ref{fig:real_sim_setup}, we constructed a digital twin simulation environment that closely mirrors the real-world setup, including the robotic arm, table surface, objects, and cameras. 
We collected 100 real-world trajectories and 500 trajectories in the simulation.
We then trained and evaluated the ACT model using different ratios of real-world to simulation data, including real-world data only, simulation data only, and mixed ratios of 100:100, 100:200, 100:300, 100:400, and 100:500. 
Notably, we did not employ any sim2real transfer techniques but instead directly combined both types of data for co-training.
Figure~\ref{fig:real_sim_results} shows the success rates of ACT in both real-world and simulation environments under different experimental settings.
Our observations revealed that increasing the proportion of simulation data improved success rates in both real-world and simulation environments, thanks to our highly accurate simulation environment that closely resembles real-world conditions.
However, we also discovered that simulation data alone is insufficient for real-world performance, with real-world data playing a crucial role. 
For instance, while the combination of 100 real-world trajectories and 500 simulation trajectories achieved a 90\% success rate in the simulated environments, using simulation data alone resulted in a dramatic decrease to a 10\% success rate in the real world. 
The primary cause of failure was the cup slipping from the gripper during rotation due to insufficient grip closure. 
This suggests that significant disparities exist between simulated and real-world physics, particularly for contact-rich tasks, indicating room for improvement in simulation fidelity.

\textbf{Performance Correlations between Real and Simulation Environments.}
We further experimented to validate the correlations between performance in the real and simulation environments of the same model
We selected five tasks, trained single-task ACT and Diffusion Policy with real-world data, and evaluated them in both the real world and the digital twin simulation environment.
Table~\ref{tab:corr} shows positive correlations of the sim-real test results. 
This aligns with findings in prior research~\citep{li24simpler} and confirms the positive correlation between simulation and real-world results.

\begin{table}[tp]
  \centering
  \caption{
    Success rates of ACT and Diffusion Policy tested in real and simulation environments.
  }
  \resizebox{0.92\columnwidth}{!}{
  \begin{tabular}{l|cc|cc}
    \toprule
    \multirow{2}{*}{\textbf{Tests in Different Environments}} &
      \multicolumn{2}{c|}{\textbf{ACT}} & \multicolumn{2}{c}{\textbf{Diffusion Policy}} \\
    & \textbf{Sim} & \textbf{Real} & \textbf{Sim} & \textbf{Real} \\        
    \midrule
    \texttt{FR-PickStrawberryInBowl}  & 3/10 & 9/10 & 0/10 & 5/10  \\
    \texttt{FR-PlaceBreadPlate}       & 2/10 & 7/10 & 3/10 & 8/10  \\
    \texttt{FR-SlideCloseDrawer}      & 2/10 & 4/10 & 7/10 & 10/10 \\
    \texttt{FR-PlacePearBowl}         & 0/10 & 4/10 & 4/10 & 10/10 \\
    \texttt{FR-PlaceBlockPlate}       & 0/10 & 3/10 & 2/10 & 7/10  \\
    \midrule
    Pearson Correlation Coefficient & \multicolumn{2}{c|}{0.83} & \multicolumn{2}{c}{0.91} \\
    \bottomrule
  \end{tabular}
  }
  \label{tab:corr}
  \vspace{-1em}
\end{table}

\section{Discussion and Future Work}

In this work, we introduce \ours, a large-scale, multi-embodiment dataset for robot manipulation. 
\ours includes four distinct embodiments, {\ntrajs} high-quality demonstrations across {\ntasks} tasks, {\nobjs} objects, and {\nskills} unique skills, collected through an intelligent data platform with a carefully designed quality assurance process.

We present quantitative analyses of \ours, highlighting its heterogeneous embodiments, diverse episode lengths, broad task coverage, and a wide range of objects drawn from five common scenarios: domestic, industrial, kitchen, office, and retail. 
We also compare \ours qualitatively with the Open X-Embodiments dataset, considering factors such as uniform settings, multiple viewpoints, and embodiment diversity. 
These analyses underscore the richness of \ours and its potential to advance research in robot manipulation.

We conduct experiments on several popular imitation learning robot models, assessing their pre-training performance and generalization capabilities on \ours. 
Our results indicate an urgent need to enhance accurate positioning and precise control in current algorithms, especially for long-horizon tasks. 
For potential investigations, we suggest that the high-quality, diverse data of \ours is especially suited—but not limited—to fostering cross-embodiment generalization, adapting imitation learning models to downstream tasks, and exploring data augmentation strategies for improved visual- and task-level generalization.

As an ongoing research project, we continue to expand \ours using standardized collection and quality assurance procedures.
Therefore, we believe \ours can serve as a ready-to-use dataset and consistently boost progress in embodied AI research.

\section{Limitations}
One limitation of \ours is the relatively simple background environments. 
While our primary focus has been on constructing a large-scale, high-quality set of robotic trajectories, we plan to investigate in the future whether incorporating more complex backgrounds can enhance the model's manipulation performance, either through data generation or further collection efforts.
Additionally, although \ours covers a broad range of robot tasks and environments, it currently lacks data from mobile manipulation scenarios. However, since two of our robotic embodiments are mobile, we plan to expand \ours to include mobile manipulation tasks in the future.
Additionally, \ours can be further enriched by adding more informative annotations such as high-level planning instructions and annotations of trajectory quality.

\section*{Acknowledgments}

This dataset and benchmark for robotic arm manipulation tasks represent a complex system engineering effort that required extensive collaboration among numerous researchers across multiple domains.
The development of this work would not have been possible without the dedication and expertise of many individuals who contributed their time and knowledge throughout various stages of the project.

We would like to extend our deepest gratitude to the following individuals for their invaluable help in this work:
Dapeng Wang, Gang Han, Haifang Huang, Jialiang Shu, Jian Xiao, Jianwei Guo, Jianyu Dong, Jiaxing Wei, Jieyu Zhang, Kai Yang, Kun Niu, Lili Chen, Manzhan Wang, Musen Zhang, Peng Guo, Qiu Cui, Shuyi Zhang, Yaowen Xu, Yijie Guo, Yingjuan Tang, Yizhang Liu, Yue Zhao, Yuheng Zhang, Zehui Liu, Zhen Hao, and Zhennan Zhang.
We also sincerely appreciate the dedication and effort of numerous contributors who assisted with data collection, quality assurance, annotation, and testing procedures.
Their collective efforts and expertise have been instrumental in making this research possible. 
We sincerely appreciate their commitment to advancing the field of robotic manipulation through this collaborative endeavor.

This work was in part supported by the National Natural Science Foundation of China (62476011).

\section*{Author Contributions}

\begin{itemize}
  \item \textbf{Project Leaders:}
    Zhengping Che and Xiaozhu Ju
  \item \textbf{Project Coordinators:}
    Kun Wu, Zhuqin Yang, Chengkai Hou, and Jiaming Liu
  \item \textbf{Data Collection and Processing:}
    Zhiyuan Xu, Guang Yang, Fei Liao, Zhen Zhao, Guangyu Li, Zhao Jin, Lecheng Wang, Kun Wu, Meng Li, and Pei Ren
  \item \textbf{Dataset Annotation:}
    Yulin Luo, Zeyu Gao, Zhenyu Wang, and Sixiang Qian
  \item \textbf{Algorithm Development:}
    \begin{itemize}
      \item ACT: Kun Wu
      \item Diffusion Policy: Kun Wu, Jilei Mao, and Xinhua Wang
      \item BAKU: Meng Li
      \item OpenVLA: Yaoxu Lyu, Xingyu Wang, Chenxuan Li, Chenyang Gu, and Yankai Fu
      \item RDT-1B: Di Wu, Jingyang He, Sixiang Chen, and Zeyu Gao
      \item CrossFormer: Shichao Fan and Xinhua Wang
    \end{itemize}
  \item \textbf{Paper Writing:}
    Chengkai Hou, Jiaming Liu, Kun Wu, Meng Li, Yinuo Zhao, Mengzhen Liu, Xinhua Wang, Zhengping Che, and Shanghang Zhang
  \item \textbf{Project Support:} 
    Zhuqin Yang, Kun Wu, Chengkai Hou, Jiaming Liu, Yinuo Zhao, Ning Liu, Xinhua Wang, Shichao Fan, Pei Ren, Qiang Zhang, Mengzhen Liu, and Pengju An
  \item \textbf{Project Advisors:}
    Jian Tang and Shanghang Zhang
\end{itemize}



\bibliographystyle{plainnat}
\bibliography{references}

\end{document}